\documentclass[10pt,twocolumn,letterpaper]{article}

\usepackage{iccv}
\usepackage{times}
\usepackage{epsfig}
\usepackage{graphicx}
\usepackage{amsmath}
\usepackage{amssymb}
\usepackage{bm}
\usepackage{subfigure}
\usepackage{algorithm}
\usepackage{algorithmic}

\usepackage[breaklinks=true,bookmarks=false]{hyperref}

\iccvfinalcopy 

\def\iccvPaperID{****} 

\begin{document}

\title{Efficient Circle-Based Camera Pose Tracking Free of PnP}

\author{Fulin Tang\\
Institute of Automation, \\
Chinese Academy of Sciences\\
{\tt\small fulin.tang@nlpr.ia.ac.cn}
\and
Yihong Wu\\
Institute of Automation, \\
Chinese Academy of Sciences\\
{\tt\small yhwu@nlpr.ia.ac.cn}
}

\maketitle

\begin{abstract}
Camera pose tracking attracts much interest both from academic and industrial communities, of which the methods based on planar markers are easy to be implemented. However, most of the existing methods need to identify multiple points in the marker images for matching to space points. Then, PnP and RANSAC are used to compute the camera pose. If cameras move fast or are far away from markers, matching is easy to generate errors and even RANSAC cannot remove incorrect matching. Then, the result by PnP cannot have good performance. To solve this problem, we design circular markers and represent 6D camera pose analytically and unifiedly as very concise forms from each of the marker by projective invariance. Afterwards, the pose is further optimized by a proposed nonlinear cost function based on a polar-n-direction geometric distance. The method is from imaged circle edges and without PnP/RANSAC, making pose tracking robust and accurate. Experimental results show that the proposed method outperforms the state of the arts in terms of noise, blur, and distance from camera to marker. Simultaneously, it can still run at about 100 FPS on a consumer computer with only CPU.
\end{abstract}

\section{Introduction}
Camera pose tracking attracts much interest both from academic and industrial communities because of its plenty of applications in virtual reality, augmented reality and robotics.

One kind of the methods for camera pose tracking is based on planar fiducial markers. They are easy to be implemented and widely used in the scenes with less textures. There are two groups of fiducial markers: markers with square/rectangle information and markers with circular information.

The methods of camera pose tracking from markers with square or rectangle information are as follows. Kato and Billinghurst \cite{kato1999marker} gave the first augmented reality system based on fiducial markers known as ARToolkit, where the used marker is a black enclosed rectangle with simple graphics or texts, as shown in (a) of Figure 1. Ababsa and Mallem \cite{ababsa2004robust} provided simple squares enclosing a rectangle. From 4 marker points, Maidi et al \cite{maidi2010performance} developed a hybrid approach that mixes an iterative method based on the extended Kalman filter and an analytical method with direct resolution of pose parameters computation. Afterwards, the research based on
planar square markers is more and more active, for example, ARTag \cite{fiala2005artag,fiala2010designing}, ARToolkitPlus \cite{wagner2010real,wagner2007artoolkitplus}, AprilTag \cite{olson2011apriltag}, AprilTag2 \cite{wang2016apriltag} and ChromaTag \cite{DeGol:ICCV:17}. Some of them are shown in Figure 1.
\begin{figure}[t]
\centering
\begin{minipage}[b]{0.3\linewidth}
  \centering
  \includegraphics[width=\linewidth]{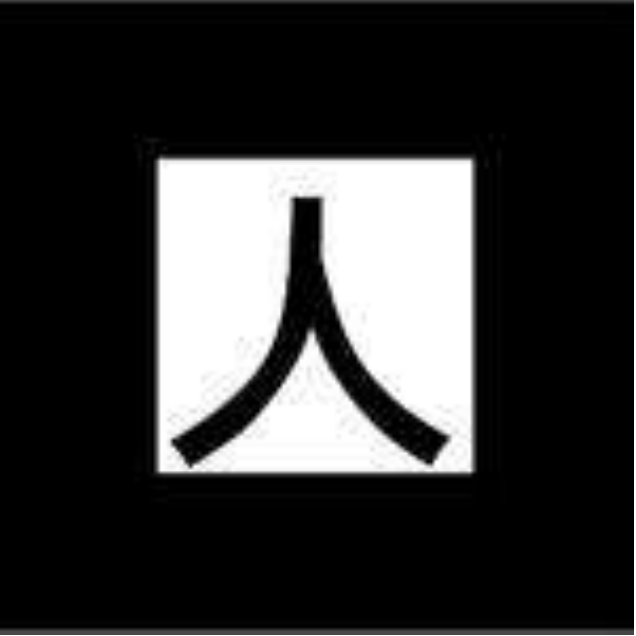}
  \centerline{(a)ARToolkit}
\end{minipage}
\begin{minipage}[b]{0.3\linewidth}
  \centering
  \includegraphics[width=\linewidth]{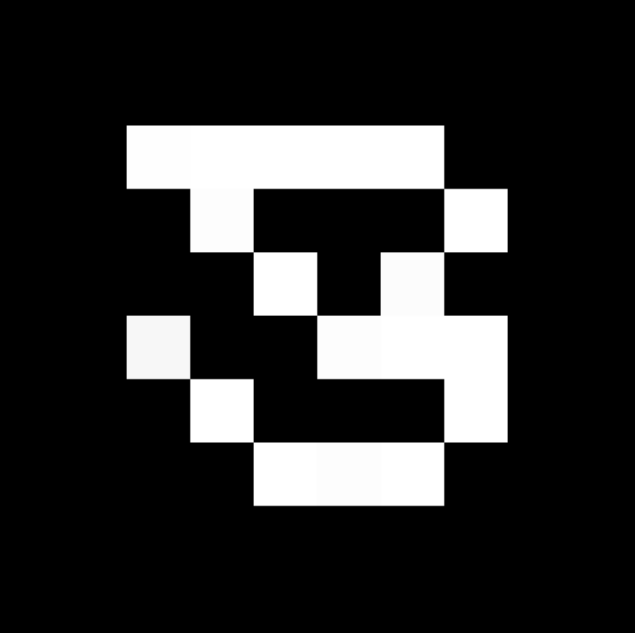}
  \centerline{(b)ARTag}
\end{minipage}
\begin{minipage}[b]{0.3\linewidth}
  \centering
  \includegraphics[width=\linewidth]{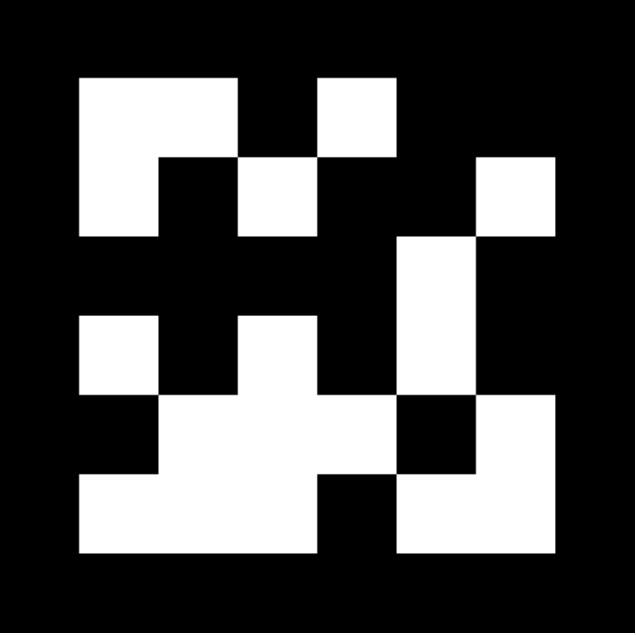}
  \centerline{(c)AprilTag}
\end{minipage}
\begin{minipage}[b]{0.3\linewidth}
  \centering
  \includegraphics[width=\linewidth]{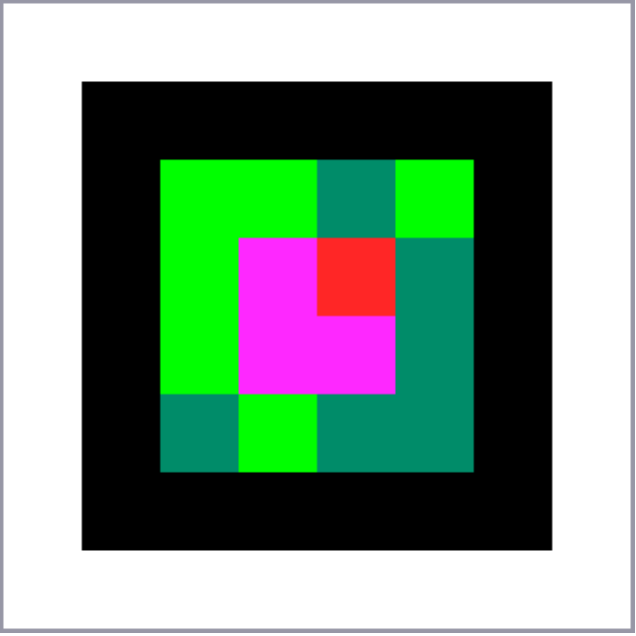}
  \centerline{(d)ChromaTag}
\end{minipage}
\begin{minipage}[b]{0.3\linewidth}
  \centering
  \includegraphics[width=\linewidth]{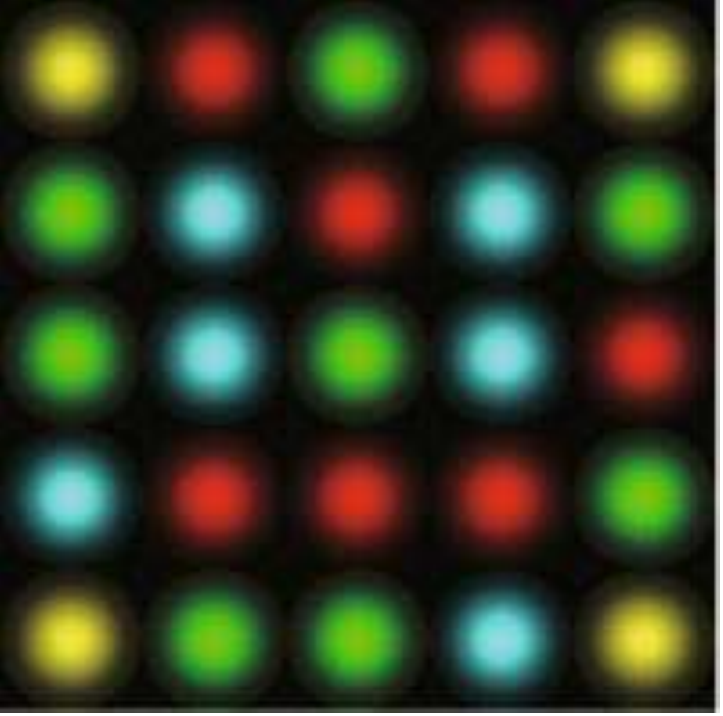}
  \centerline{(e)Mono-spectrum}
\end{minipage}
\begin{minipage}[b]{0.3\linewidth}
  \centering
  \includegraphics[width=\linewidth]{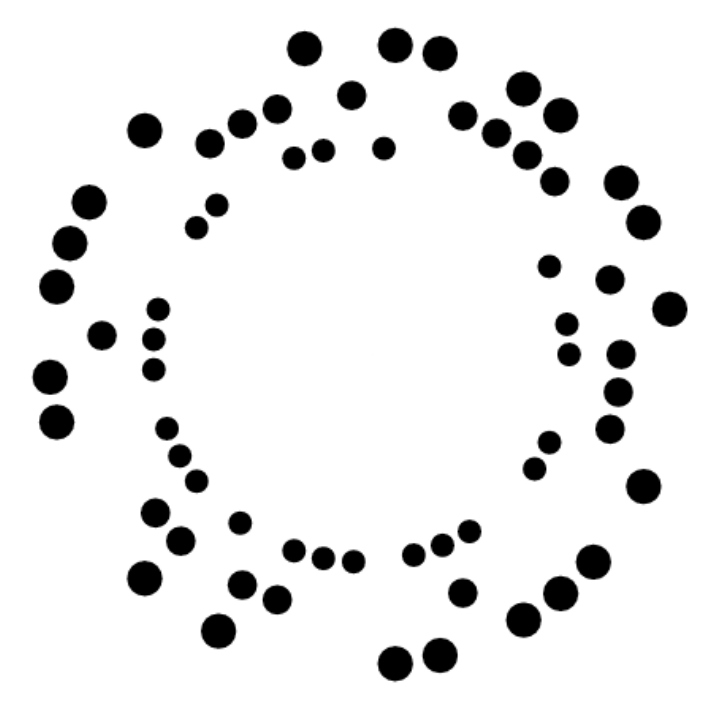}
  \centerline{(f)RUNETag}
\end{minipage}
\caption{Popular planar fiducial markers. (a)ARToolkit with arbitrary pixel patterns inside the black border; (b)ARTag and (c)AprilTag using binary digital codes; (d)ChromaTag using color information; (e)Mono-spectrum using low frequency components with coloured dots on a black square background; (f)RUNETag with dots lying on circles.}
\label{fig:long}
\label{fig:onecol}
\end{figure}
ARTag is a black enclosed rectangle with black and white blocks where digital coding theory is used to get a low false positive and inter-marker confusing rate. It has 2002 markers, a larger library and no need of pattern files, with improved performance to ARToolkit in identification and verification. ARToolkitPlus is an open source marker tracking library designed as a successor to the extremely popular open source library ARToolkit. Compared to ARToolkit, the performance of ARToolkitPlus is improved and it can be implemented with real time on mobile device. AprilTag is a stronger digital coding system and greater robustness to occlusion, warping, and lens distortion. AprilTag2 proposed a completely redesigned tag detector that improves robustness and efficiency of AprilTag. ChromaTag uses the information of opponent colors to limit and quickly reject initial false detections and grayscale.

Markers with circular information have two categories: one category only gives the markers and the corresponding detections, where no camera pose computation is provided. Another category gives markers and camera pose computations. Methods of the first category are as follows. In 1992, Gatrell et al \cite{gatrell1992robust} gave a concentric contrasting circle as well as its extraction by computing centroids.  Later, the concentric circle approach was improved by adding additional color and scale information by Cho et al \cite{neumannmulti}. Ring information was considered into the marker by Knyaz and Sibiryakov \cite{knyaz1998development}. Naimark and Foxlin \cite{naimark2002circular} developed a more general marker generation method, which encodes the bar code into the black circular region to produce more markers. Toyoura et al \cite{toyoura2014mono} designed mono-spectrum marker, as shown in (e) of Figure 1, where low frequency components in the form of coloured ¡°degraded¡± dots lying on a black square background are used and the computation needs GPU. Both of Prasad et al \cite{prasad2015motion} and Calvet et al \cite{calvet2016detection} gave extractions for images of concentric circles that are robust to blur. Methods of the second category with circular information are as follows. A square with four circles at its corners was proposed by Claus and Fitzgibbon \cite{claus2005reliable}. Calvet et al \cite{calvet2012camera} computed rotations from concentric circles by factorization. Bergamasco et al \cite{bergamasco2016accurate} designed a set of circular high-contrast dots arranged into concentric layers called RUNETag as shown in (f) of Figure 1. PiTag \cite{bergamasco2013pi} also used small circular dots but they are arranged in rectangles. Chen et al \cite{chen2004camera} computed camera pose from two coplanar separate circles.

Most of the above methods to compute camera poses need PnP based on matching between image and space points. If the used camera moves fast or is far away from markers, the matching is easy to be wrong. Even though RANSAC is employed, the performance is not good because usually the number of points in markers is small.
\cite{calvet2012camera} and \cite{chen2004camera} don't need matching, but the computations for camera pose are complex and there is only one kind of configuration of circles. Moreover, the computation is only for a single image in \cite{chen2004camera} that cannot be used for videos due to lacking a unified world coordinate system. Both of \cite{calvet2012camera} and \cite{chen2004camera} didn't report the computation speed.

In this paper, we design circular markers to compute 6D camera poses from videos. The camera poses are unified into a global world coordinate system. PnP and RANSAC are not needed. We use projective invariance to identify the origin and the orthogonal coordinate axes of world coordinate system from images. The camera pose is analytically and unifiedly represented as concise forms. Furthermore, we establish a nonlinear cost function based on a polar-n-direction geometric distance \cite{wu2019efficient} to optimize the analytical pose. Due to the concise representations, the proposed nonlinear optimization can always converge quickly. Therefore our method can run in real time with about 100 FPS on a consumer computer with only CPU. Simultaneously, it is robust to noise, blur, and distance from camera to marker with high accuracy. Experiments prove that our method outperforms the state of the arts.

\begin{figure}[t]
\begin{center}
   \includegraphics[width=0.5\linewidth]{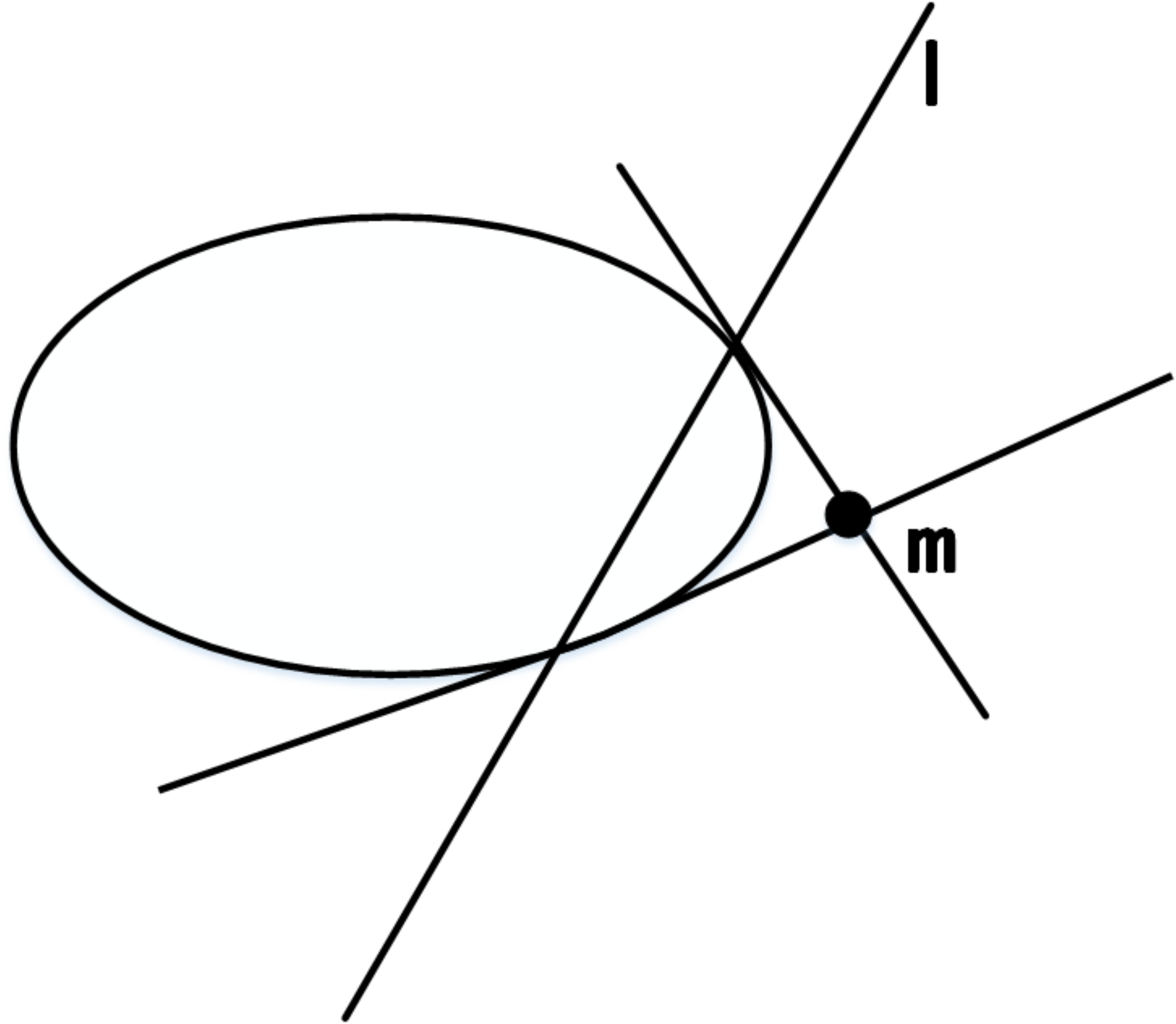}
\end{center}
   \caption{A point $\bm{m}$ and its polar line $\bm{l = Cm}$ related to a conic $\bm{C}$}
\label{fig:long}
\label{fig:onecol}
\end{figure}
\section{Preliminaries}
A bold letter denotes a vector or a matrix. $\approx$ denotes equality up to a scale. Homogeneous coordinates of a 2D planar point is in the form ${(x, y, w)}^\mathrm{T}$. Cross ratio is kept invariant under a projective transformation. Each circle and the line at infinity of the circle supporting plane intersect at a pair of circular points. Two orthogonal lines on this plane intersect with the line of infinity at two points. The two points are harmonic, i.e. the cross ratio of the two points and the two circular points is $-1$, and are conjugate points with respect to the circle. Also, the two orthogonal lines are conjugate with respect to the circle. In fact, any two orthogonal lines are conjugate with respect to any a circle on the same plane.

The locus of planar points with homogeneous coordinates ${(x, y, w)}^\mathrm{T}$ that satisfies:
\begin{equation}
\begin{split}
	\begin{pmatrix}
		x & y & w
	\end{pmatrix}
	\begin{pmatrix}
		a & d & e \\
		d & b & f \\
		e & f & c
	\end{pmatrix}
	\begin{pmatrix}
		x \\
		y \\
		w
	\end{pmatrix}
	= 0
\end{split}
\end{equation}
is a conic. We denote
$
\left(
\begin{array}{ccc}
a & d & e \\
d & b & f \\
e & f & c
\end{array}
\right)
$
as $\bm{C}$ that can represent this conic. A conic is an element in a projective plane. For example, a circle in space is projected as a conic by a perspective camera.

Given a point $\bm{m}$ in 2D homogeneous coordinates, let $\bm{Cm}$ represent a line (i.e. $\bm{Cm}$ is as line coordinate). Then $\bm{m}$ and $\bm{Cm}$ are of polarity relationship related to $\bm{C}$. The relationship is invariant under a projective transformation. $\bm{Cm}$ is called the polar line of $\bm{m}$ related to $\bm{C}$ and inversely $\bm{m}$  is called the pole of $\bm{Cm}$. Figure 2 shows the polarity relationship of a point $\bm{m}$ and its polar line $\bm{l = Cm}$.

\section{Designed Markers}
The proposed markers are based on circles. Six kinds of them are shown in Figure 3 respectively. They are simple and common in our life. Also, it is convenient to make them.
\begin{figure}[t]
\centering
\begin{minipage}[b]{0.45\linewidth}
  \centering
  \includegraphics[width=\linewidth]{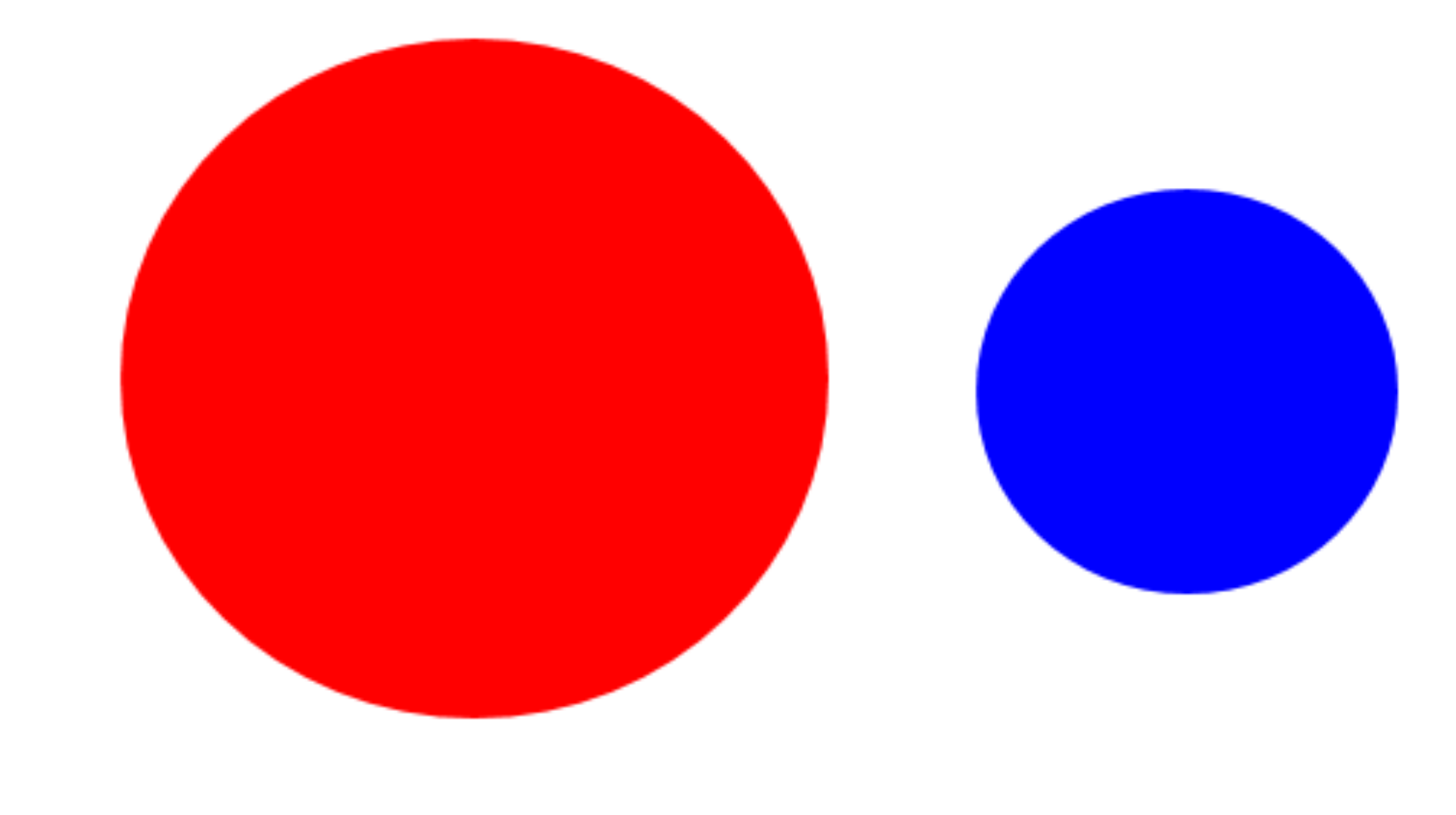}
  \centerline{(a)}
\end{minipage}
\begin{minipage}[b]{0.45\linewidth}
  \centering
  \includegraphics[width=\linewidth]{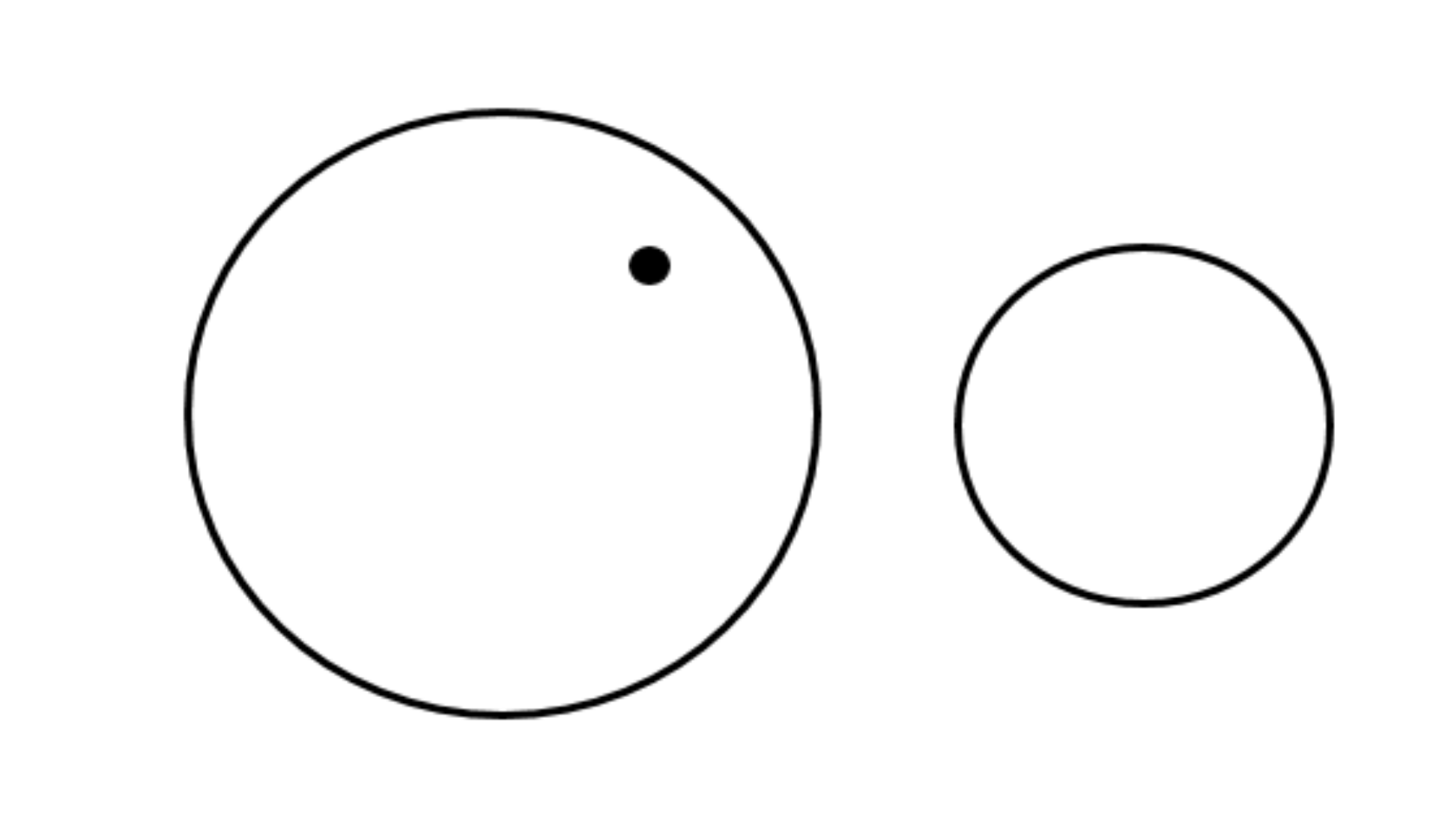}
  \centerline{(b)}
\end{minipage}
\begin{minipage}[b]{0.45\linewidth}
  \centering
  \includegraphics[width=\linewidth]{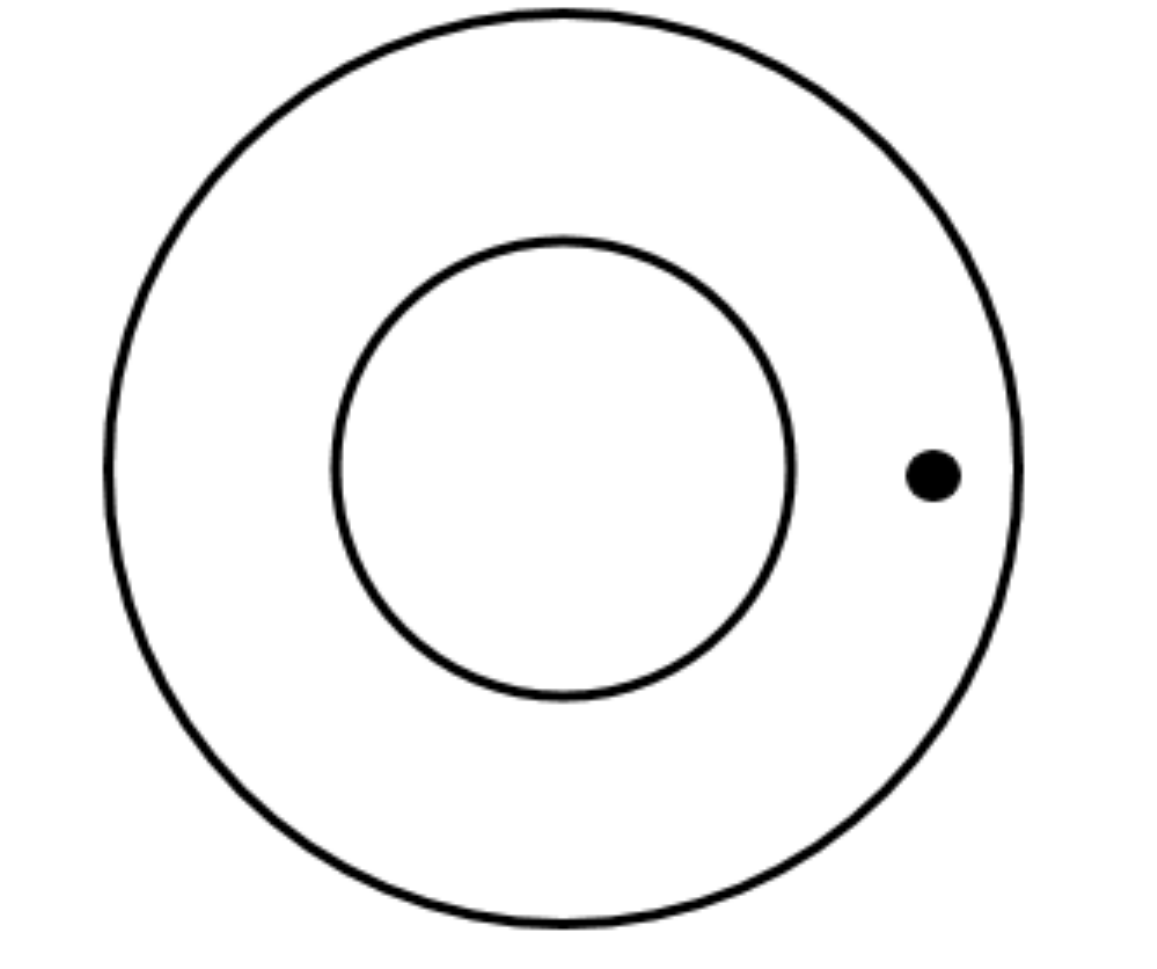}
  \centerline{(c)}
\end{minipage}
\begin{minipage}[b]{0.45\linewidth}
  \centering
  \includegraphics[width=\linewidth]{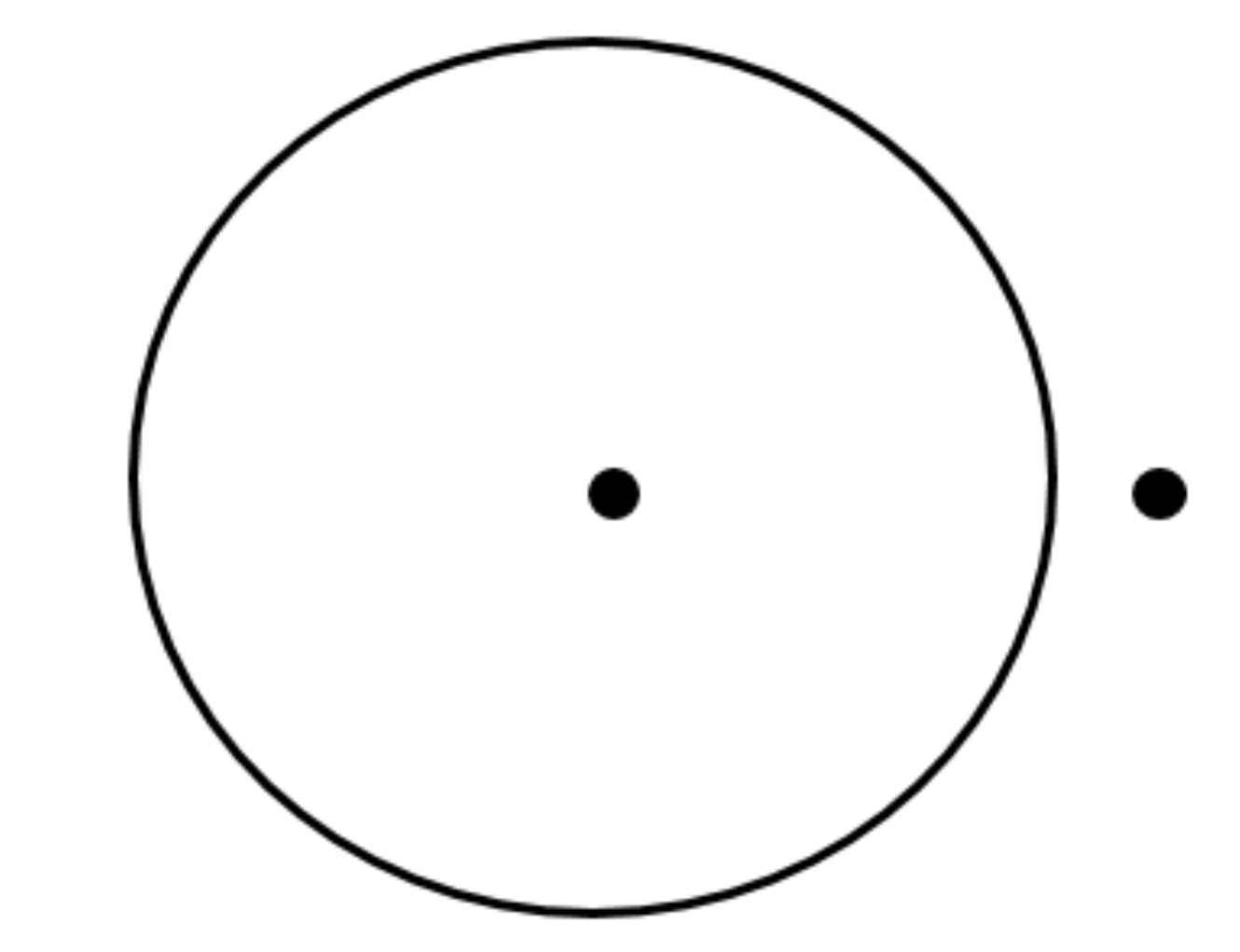}
  \centerline{(d)}
\end{minipage}
\begin{minipage}[b]{0.45\linewidth}
  \centering
  \includegraphics[width=\linewidth]{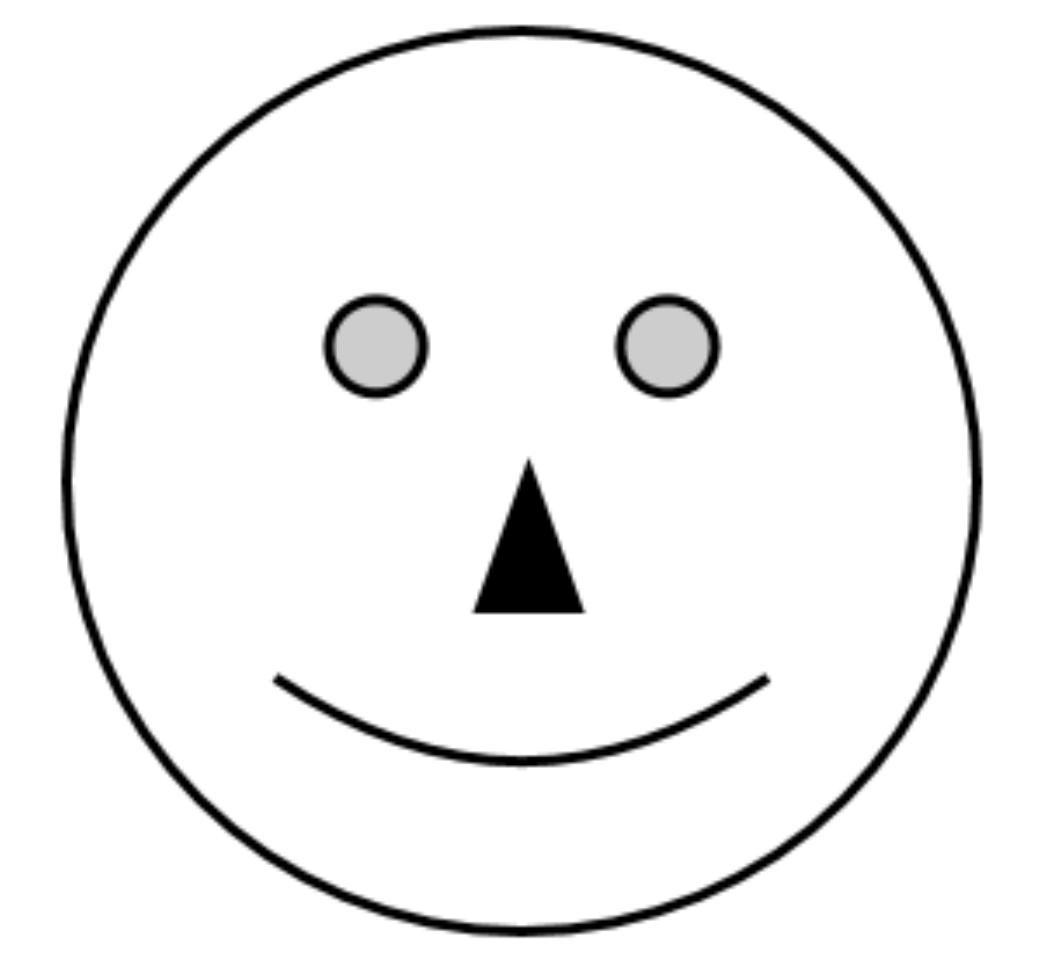}
  \centerline{(e)}
\end{minipage}
\begin{minipage}[b]{0.45\linewidth}
  \centering
  \includegraphics[width=\linewidth]{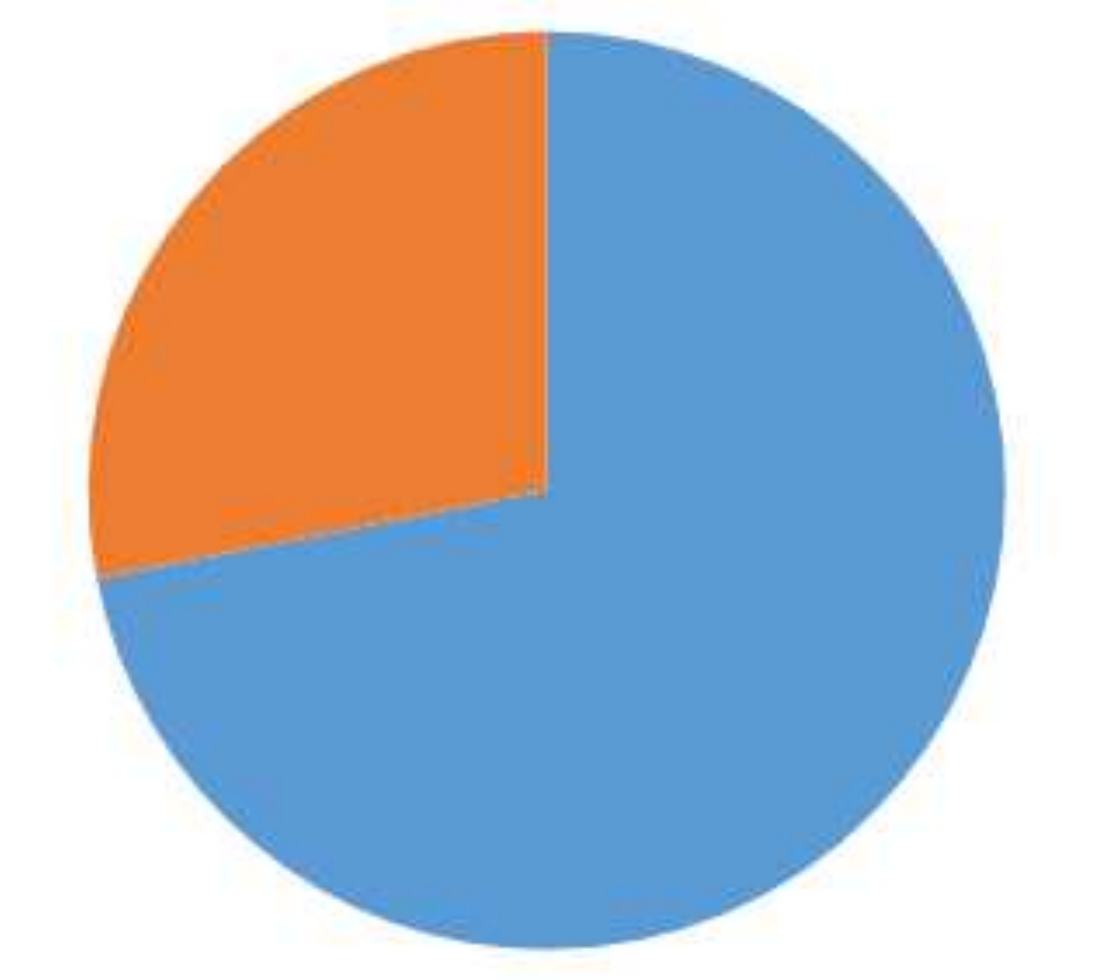}
  \centerline{(f)}
\end{minipage}
\caption{The designed markers}
\label{fig:long}
\label{fig:onecol}
\end{figure}
The first three kinds of markers consist of two circles. In (a), the marker consists of two circles with different colors, for example red and blue. The different colors are used to distinguish the two circles. In (b) and (c), besides two circles, there is one point. The point is used to localize a world coordinate axis. In (d), (e) and (f), the markers consist of one circle, the circle center, and one direction for a world coordinate axis.

For each group in Figure 3, we extract two points $\bm{m_0}$ and $\bm{m_1}$ before localizing the camera. In (d), (e) and (f), $\bm{m_0}$ is the image of the circle center that can be extracted from the images. In (d), $\bm{m_0}$ is the imaged point inside the circle and $\bm{m_1}$ is the imaged point outside the circle. In (e), $\bm{m_0}$ is recognized as the imaged nose tip, and $\bm{m_1}$ is any of the imaged mouth tips. Once $\bm{m_1}$ is determined, it will be fixed as the mouth tip in the next video frames. In (f), $\bm{m_0}$ is the intersection of the two radii edges, and $\bm{m_1}$ is any of the intersections of the radii with the circle edge. Similarly, once $\bm{m_1}$ is determined, it will be fixed as the location in the next frame. In (a), (b) and (c), $\bm{m_0}$ is one of the imaged circle centers that can not be extracted directly but can be computed out by the imaged circles. In (a), we let $\bm{m_0}$ and $\bm{m_1}$ be the two imaged circle centers. Once $\bm{m_0}$ and $\bm{m_1}$ are determined, they are fixed as that imaged circle center respectively in the next video frames. In (b), there is a feature point in one circle. We let $\bm{m_1}$ be the image of this feature point. And, let $\bm{m_0}$ be the imaged center of the circle containing this feature point. In (c), $\bm{m_0}$ is the imaged circle center of the concentric circles and $\bm{m_1}$ is the image of the feature point between the two circles. How to compute $\bm{m_0}$ and $\bm{m_1}$ in (a)? And how to compute $\bm{m_0}$ in (b) and (c)? We give the method as follows.

A circle is imaged as a conic by a perspective camera. The conic from our markers is fitted by a polar-n-direction geometric distance method \cite{wu2019efficient}. Then, the representation matrix defined in (1) is obtained. Furthermore based on the quasi-affine invariance proposed in \cite{wu2004camera}, the image of line at infinity $\bm{l_\infty}$ from (a), (b) and (c) are computed respectively. According to the invariance of polarity relationship between $\bm{l_\infty}$ and circle centers, the imaged centers of the two circles are computed as $\bm{m_0}$ and $\bm{m_1}$, where we write the coordinates of $\bm{m_0}$ as $(u_0, v_0, 1)^\mathrm{T}$ and of $\bm{m_1}$ as $(u_1, v_1, 1)^\mathrm{T}$. The coordinates will be used later for computing camera pose in the next section. Now, we get $\bm{m_0}$, $\bm{m_1}$ and $\bm{l_\infty}$ for each group of (a) (b) (c). Also we have gotten $\bm{m_0}$ and $\bm{m_1}$ by extracting for (d), (e) and (f). Similarly, by the invariance of polarity relationship between $\bm{l_\infty}$ and circle centers, we can get $\bm{l_\infty}$ as :
\begin{equation}
\bm{l_\infty} = \bm{C}\bm{m_0}
\end{equation}
for (d) (e) and (f).

Denote the transformation from the space points under the world coordinate system to the space points under a camera coordinate system as $\bm{(R,t)}$, where $\bm{R}$ is a 3x3 rotation matrix with $\bm{R = (r_1, r_2, r_3)}$, ${\bm{r_1} = (r_{11}, r_{21}, r_{31})}^\mathrm{T}$, ${\bm{r_2} = (r_{12}, r_{22}, r_{32})}^\mathrm{T}$, ${\bm{r_3} = (r_{13}, r_{23}, r_{33})}^\mathrm{T}$ and $\bm{t}$ is a 3-vector. $\bm{(R,t)}$ is the camera pose that can be analytically represented by $\bm{m_0}$, $\bm{m_1}$ and $\bm{l_\infty}$ from each image into a unified world coordinate system. The process is given in the following Section 4.

\section{Camera Pose Computation}
Before calculating camera pose, the camera intrinsic parameter matrix $\bm{K}$ is calibrated and the image is transformed by $\bm{K}^{-1}$. Then, we establish the world coordinate system as: the space point of $\bm{m_0}$ is as the origin, the straight line from the space point of $\bm{m_0}$ to the space point of $\bm{m_1}$ as $X$ axis, the line vertical to $X$ axis on the circle plane as $Y$ axis, and then $Z$ axis is vertical to $X$ axis and $Y$ axis and facing to the camera. For example, the world coordinate system of marker (a) in Figure 3 is shown in Figure 4. Further, we need to know the distance between the space point of $\bm{m_0}$  and the space point of $\bm{m_1}$. The distance is denoted as $L_x$. By the established world coordinate system, the space point of $\bm{m_1}$ is $(L_x, 0, 0, 1)^{T}$.
\begin{figure}[t]
\begin{center}
   \includegraphics[width=1.0\linewidth]{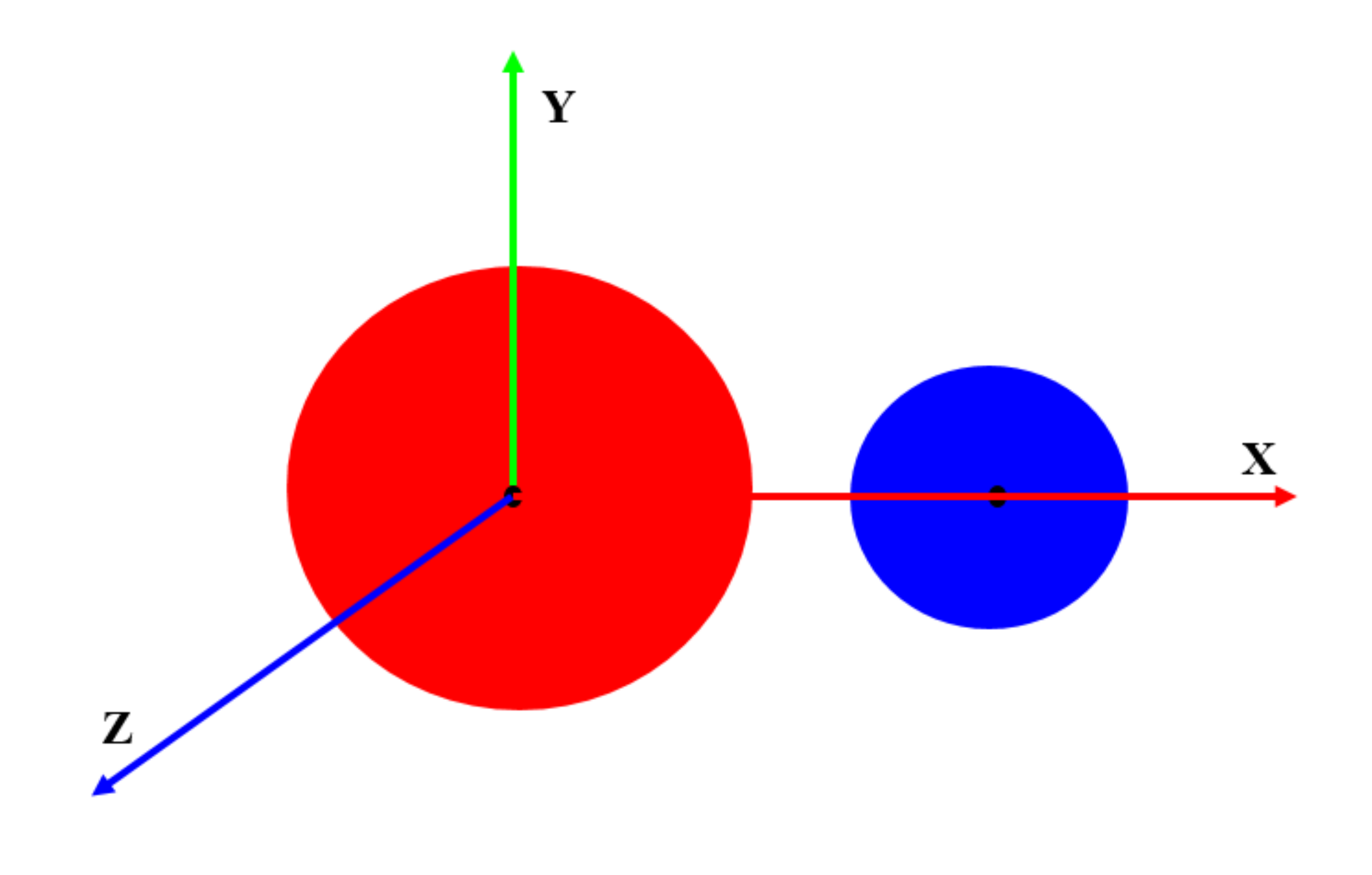}
\end{center}
   \caption{World coordinate system for marker (a) in Figure 3}
\label{fig:long}
\label{fig:onecol}
\end{figure}

Since the space point of $\bm{m_0}$ is set as the origin of the world coordinate system, we have:
\begin{equation}
    \bm{m_0}
    \approx
	\begin{pmatrix}
		\bm{r_1} & \bm{r_2} & \bm{r_3} & \bm{t}
	\end{pmatrix}
	\begin{pmatrix}
		0 \\
		0 \\
		0 \\
        1
	\end{pmatrix}.
\end{equation}
By expanding (3), we get:
\begin{equation}
\bm{t} = s_0 *  \bm{m_0},
\end{equation}
where $s_0$ is a scale that needs to solve.

According to imaging process of the infinite point in $X$ axis, we have:
\begin{equation}
    (\bm{m_0} \times \bm{m_1}) \times \bm{l_\infty}
    \approx
    \begin{pmatrix}
		\bm{r_1} & \bm{r_2} & \bm{r_3} & \bm{t}
	\end{pmatrix}
	\begin{pmatrix}
		1 \\
		0 \\
		0 \\
        0
	\end{pmatrix}.
\end{equation}
By expanding (5) and unity norm of ${\bm r_1}$, we get:
\begin{equation}
    \bm{r_1} =
    \frac{(\bm{m_0} \times \bm{m_1}) \times \bm{l_\infty}}{\|{(\bm{m_0} \times \bm{m_1}) \times \bm{l_\infty}}\|}.
\end{equation}
Up to now, $\bm{r_1}$ is solved.

The space point of $\bm{m_1}$ is $(L_x, 0, 0, 1)^\mathrm{T}$. Then:
\begin{equation}
    \bm{m_1}
    \approx
	\begin{pmatrix}
		\bm{r_1} & \bm{r_2} & \bm{r_3} & \bm{t}
	\end{pmatrix}
	\begin{pmatrix}
		L_x \\
		0 \\
		0 \\
        1
	\end{pmatrix}.
\end{equation}
Expanding (7) and substituting (4) into the result give:
\begin{equation}
s_1*\bm{m_1} = L_x*\bm{r_1} + s_0*\bm{m_0},
\end{equation}
where $s_1$ is another scale we haven't known. Substituting (6) into (8) and solving (8) for $s_0$ and $s_1$ yield:
\begin{equation}
s_0 = \frac{L_x(r_{11} - r_{31}u_1)}{u_1 - u_0} = \frac{L_x(r_{21} - r_{31}v_1)}{v_1 - v_0},
\end{equation}
\begin{equation}
s_1 = \frac{L_x(r_{11} - r_{31}u_0)}{u_1 - u_0} = \frac{L_x(r_{21} - r_{31}v_0)}{v_1 - v_0}.
\end{equation}
Then, $\bm{t}$ is determined uniquely by (4) and (9).

According to perspective process, there is :
\begin{equation}
    \bm{l_\infty}
    \approx
    \bm{r_3}.
\end{equation}
So we obtain:
\begin{equation}
\bm{r_3} = s3 * \frac{ \bm{l_\infty}}{\|\bm{l_\infty}\|},
\end{equation}
where $s_3 = \pm 1$, which can be determined uniquely by camera facing direction.

According to $\bm{r_1}$ and $\bm{r_3}$, $\bm{r_2}$ is given:
\begin{equation}
\bm{r_2} = \bm{r_3} \times \bm{r_1}.
\end{equation}

In above, the camera pose $\bm{R = (r_1, r_2, r_3)}$, $\bm{t}$ are obtained. They are analytically represented as(4), (6), (12), (13) with the solved $s_0$, $s_1$ and $s_3$.

\section{Camera Pose Tracking with Nonlinear Optimization}
When a camera captures videos containing the proposed markers in Figure 3, 6D pose of the camera can be computed in real time and online. In order to get more accurate and robust result, we further optimize the analytical solutions of $\bm{R = (r_1, r_2, r_3)}$ and $\bm{t}$ given in Section 4 with a proposed nonlinear cost function.

Let $\bm{P}$ be the space points on circles. Then, according to camera projection model, its image point is:
\begin{equation}
     \bm{p}
     \approx
     \bm{\pi}(\bm{P}) =
     \bm{K}
	\begin{pmatrix}
		\bm{R} & \bm{t}
	\end{pmatrix}
    \bm{P},
\end{equation}
where $\bm{\pi}$ is denoted as the camera model.
If $\bm{K(R,t)}$ are accurate and the fitted conics $\bm{C}$ on the image plane have no noise, the distance \cite{wu2019efficient} between $\bm{p}$ and $\bm{C}$, denoted as $d^2(\bm{p}, \bm{C})$, is zero. Otherwise, $d^2(\bm{p}, \bm{C})$ isn't. Therefore, we go to minimize the following function:
\begin{equation}
{\sum}d^2(\bm{p}, \bm{C}) = {\sum}_{\bm{\mathcal{Y}}}d^2_{1}(\bm{p}, \bm{C}) + \frac{1}{4}{\sum}_{\bm{\mathcal{N}}}d^2_{2}(\bm{p}, \bm{C}),
\end{equation}
where
\begin{equation}
\begin{split}
	&d^2_{1} \left( \bm{p},\, \bm{C} \right)
	= \\
	&\frac
	{
		{
		\left(
			{\bm{p}}^{T} \bm{Cp}
		\right)
		}^2
	}
	{
		{
		\left(
			1
			+
			\sqrt
			{
				\frac
				{
					{
					\left(
						{\bm{p}}^{T} \bm{Gp}
					\right)
					}^2
					-
					\left(
						{\bm{p}}^{T} \bm{Cp}
					\right)	
					\left(
						{\bm{p}}^{T} \bm{Wp}
					\right)										
				}
				{
					{
					\left(
						{\bm{p}}^{T} \bm{Gp}
					\right)
					}^2
				}
			}
		\right)	
		}^2
		\left(
			{\bm{p}}^{T} \bm{Gp}
		\right)
	},
\end{split}
\end{equation}
\begin{equation}
d^2_{2}( \bm{p}, \bm{C}) = \frac{ {( {\bm{p}}^{T} \bm{C} {\bm{p}}) }^{2}}
{ {\bm{p}}^{T} \bm{G} {\bm{p}} },
\end{equation}
with
$
\bm{\bar{C}} =
\begin{pmatrix}
a & d & e \\
d & b & f \\
0 & 0 & 0
\end{pmatrix},
$
$
\bm{G}=\bm{C \bar{C}}$
and
$\bm{W} = {\left( \bm{\bar{C}} \right)}^T \bm{C} \bm{\bar{C}}$
, and \bm{\mathcal{Y}} means:
\begin{equation}
	{\left( {\bm{p}}^{T} \bm{G} {\bm{p}} \right)}^{2}
	\geq
	\left( {\bm{p}}^{T} \bm{C} {\bm{p}} \right)
	\left( {\bm{p}}^{T} \bm{W} {\bm{p}} \right).
\end{equation}
If (18) is not satisfied, we use \bm{\mathcal{N}} to represent it.
By substituting (14) into (15), the nonlinear cost function is given as follows:
\begin{equation}
\begin{split}
\mathop{\arg\min}_{\bm{R}, \bm{t}}({\sum}_
			{
			 \bm{\mathcal{Y}}
			}
			d^2_{1} \left( \bm{K}
	                   \begin{pmatrix}
		               \bm{R} & \bm{t}
	                   \end{pmatrix}
                       \bm{P},\, \bm{C} \right)
            + \\
            \frac{1}{4}{\sum}_{\bm{\mathcal{N}}}
            d^2_{2} \left( \bm{K}
	                   \begin{pmatrix}
		               \bm{R} & \bm{t}
	                   \end{pmatrix}
                       \bm{P},\, \bm{C} \right))
\end{split}
\end{equation}
We take the analytical solution in Section 4 as initial values and use Levenberg-Marquardt iteration method to optimize the initial camera pose. Ceres solver \cite{agarwal2012ceres} is used to implement it. Since the analytical solution is accurate, the nonlinear optimization can always converge quickly.

Finally, a camera pose tracking algorithm is summarized as follows.

\noindent\rule{\columnwidth}{1pt}
\textbf{Algorithm}: \leftline{Camera pose tracking}
\noindent\rule{\columnwidth}{1pt}
\smallbreak
	\begin{algorithmic}[1] \label{alg::this}
		\REQUIRE Videos or image sequence.
		\ENSURE Camera pose $\bm{R, t}$ for each frame.
		\STATE Extract edges and corner points for the image.
        \STATE Make conic fitting from the edges. Compute or find $\bm{m_0}$, $\bm{m_1}$ and $\bm{l_\infty}$ by the method in Section 3.
        \STATE Compute the camera initial pose $\bm{R}$, $\bm{t}$ by the method in Section 4.
        \STATE Minimize (19) by Levenberg-Marquardt iteration to refine camera pose.
        \STATE For the next frame, predict edges and the corner points. Then repeat the above step 1-4.
	\end{algorithmic}
\noindent\rule[1mm]{\columnwidth}{1pt}

\begin{figure}[t]
\centering
\begin{minipage}[b]{0.21\linewidth}
  \centering
  \framebox{\includegraphics[width=\linewidth]{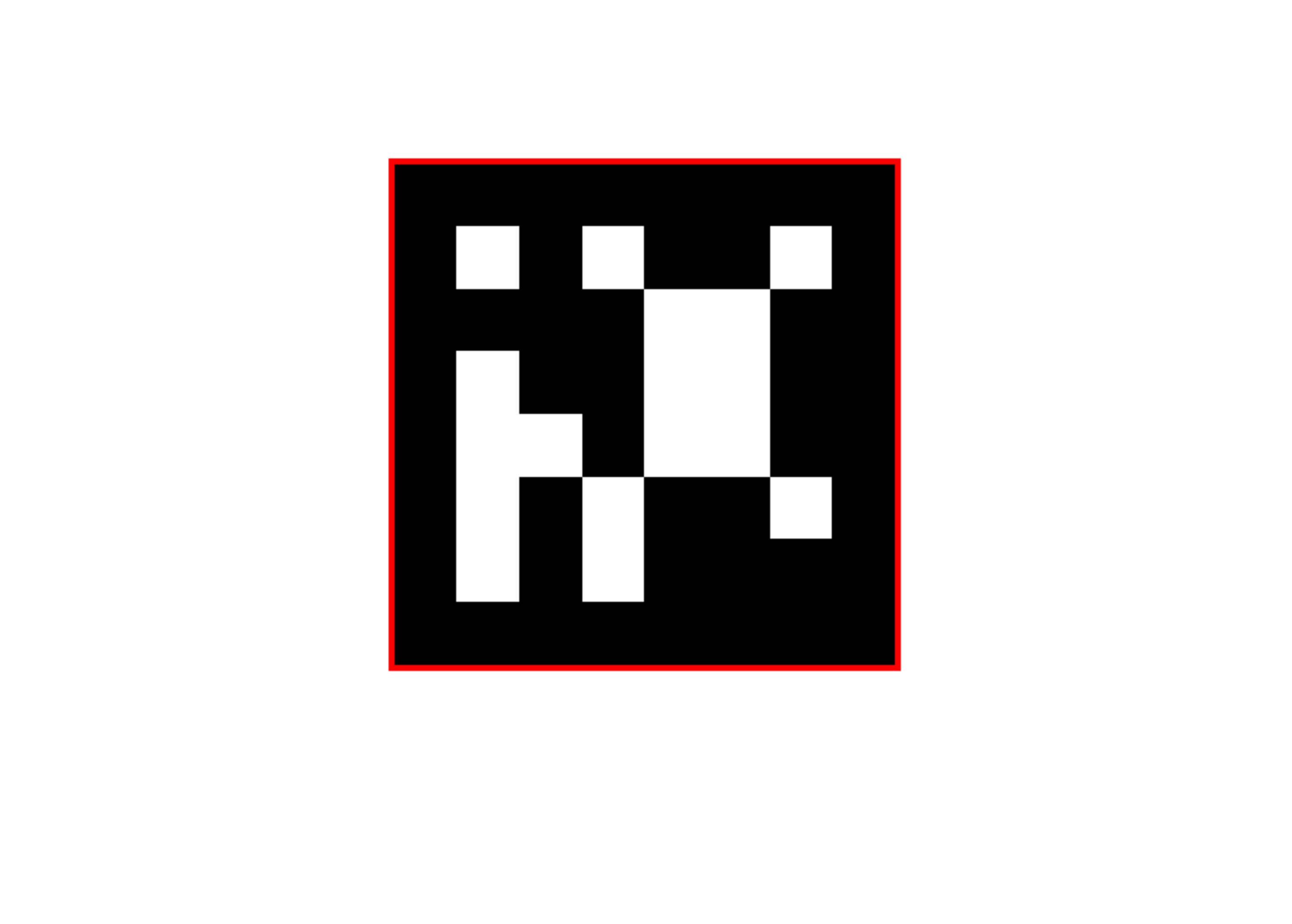}}
  \centerline{(a)}
\end{minipage}
\quad
\begin{minipage}[b]{0.21\linewidth}
  \centering
  \framebox{\includegraphics[width=\linewidth]{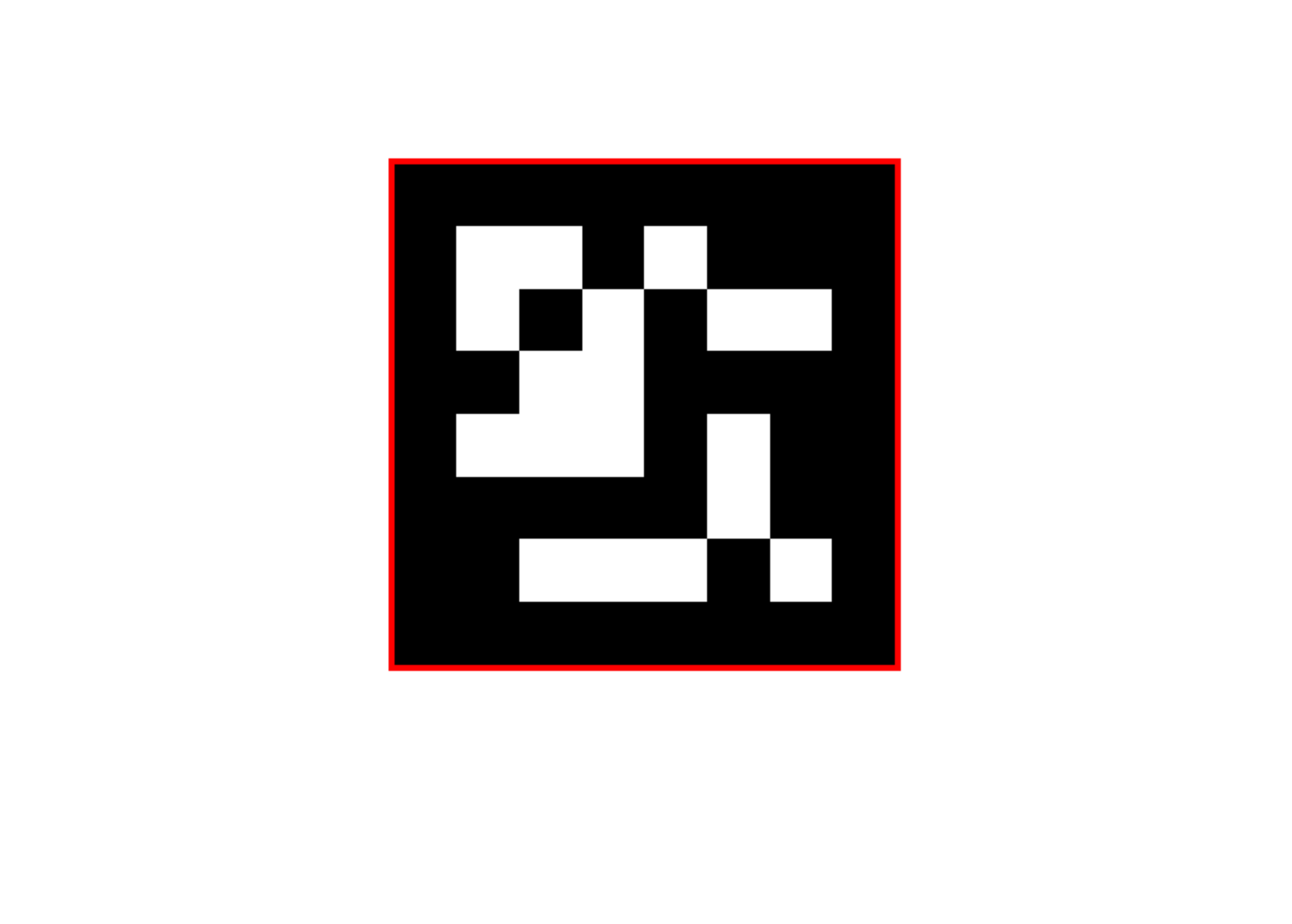}}
  \centerline{(b)}
\end{minipage}
\quad
\begin{minipage}[b]{0.21\linewidth}
  \centering
  \framebox{\includegraphics[width=\linewidth]{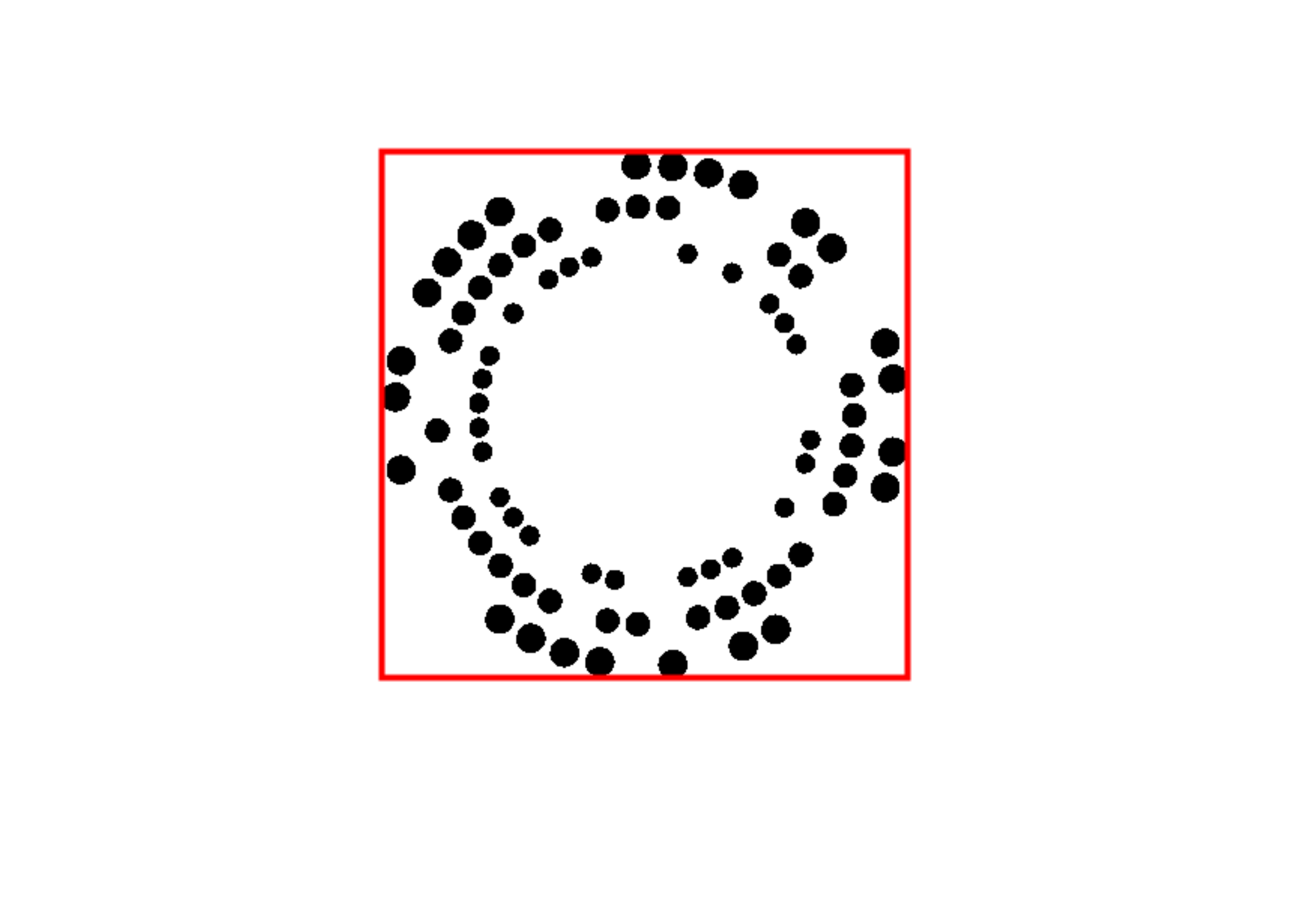}}
  \centerline{(c)}
\end{minipage}
\quad
\begin{minipage}[b]{0.21\linewidth}
  \centering
  \framebox{\includegraphics[width=\linewidth]{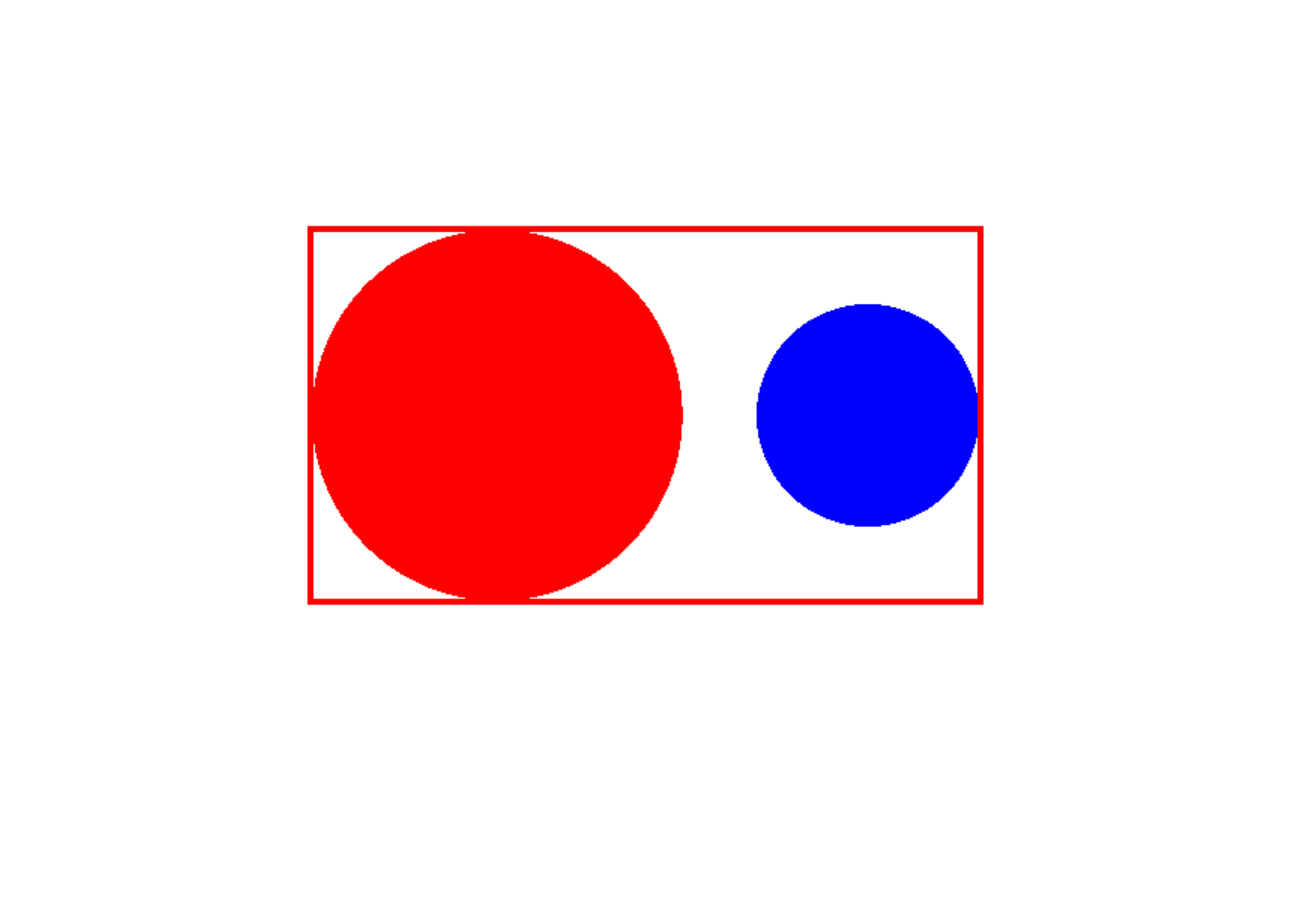}}
  \centerline{(d)}
\end{minipage}
\caption{Markers for experimental comparison. (a) is ARToolkitPlus. (b) is AprilTag2. (c) is RUNETag. (d) is the marker of our method. The area of red box in (a), (b), (c) and (d) represents the image area. The same image area for different markers is set.}
\label{fig:long}
\label{fig:onecol}
\end{figure}
\begin{figure}[t]
\centering
\begin{minipage}[b]{0.45\linewidth}  
  \centering
  \includegraphics[width=\linewidth]{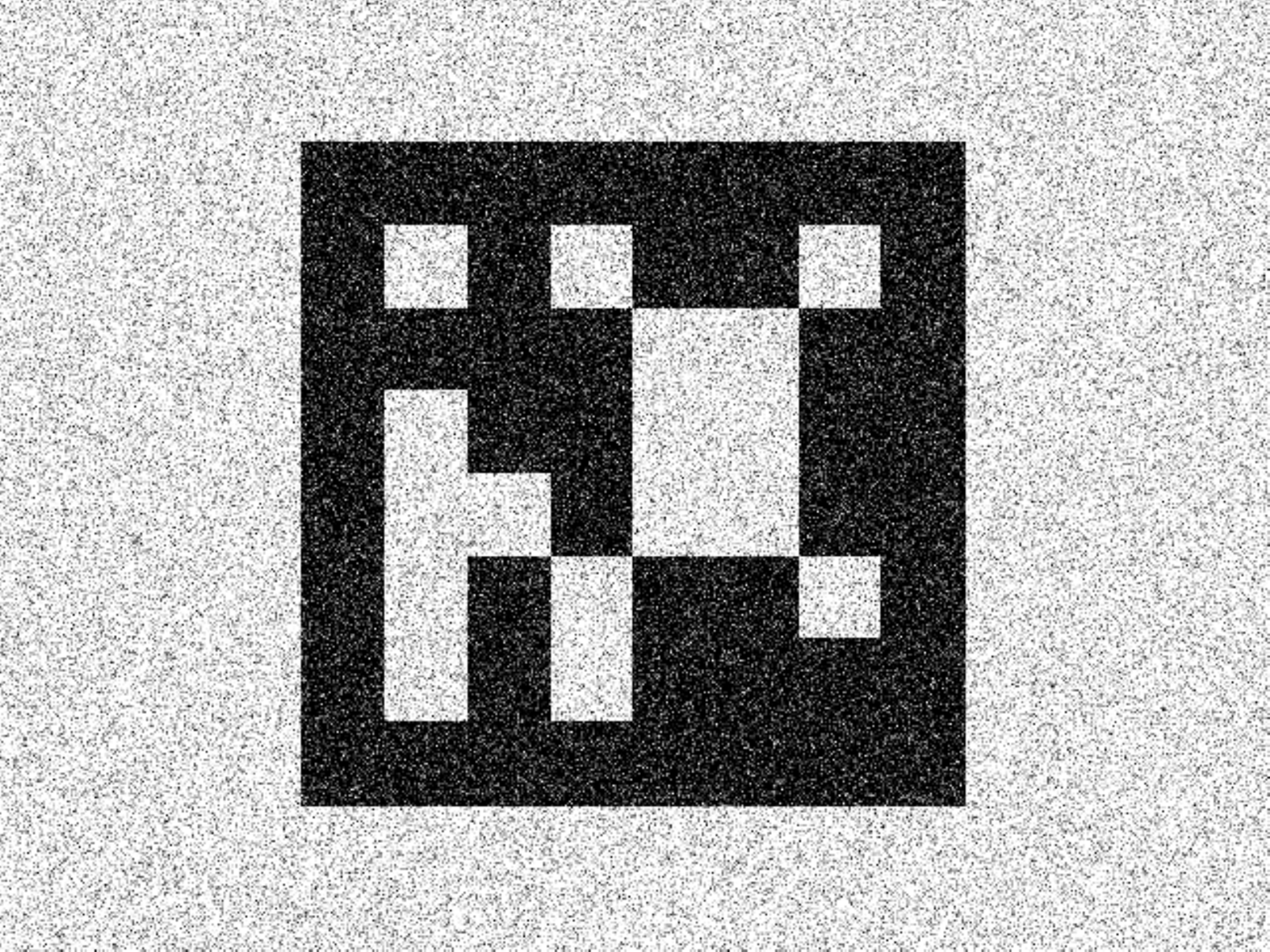}
  \centerline{(a)}
\end{minipage}
\begin{minipage}[b]{0.45\linewidth} 
  \centering
  \includegraphics[width=\linewidth]{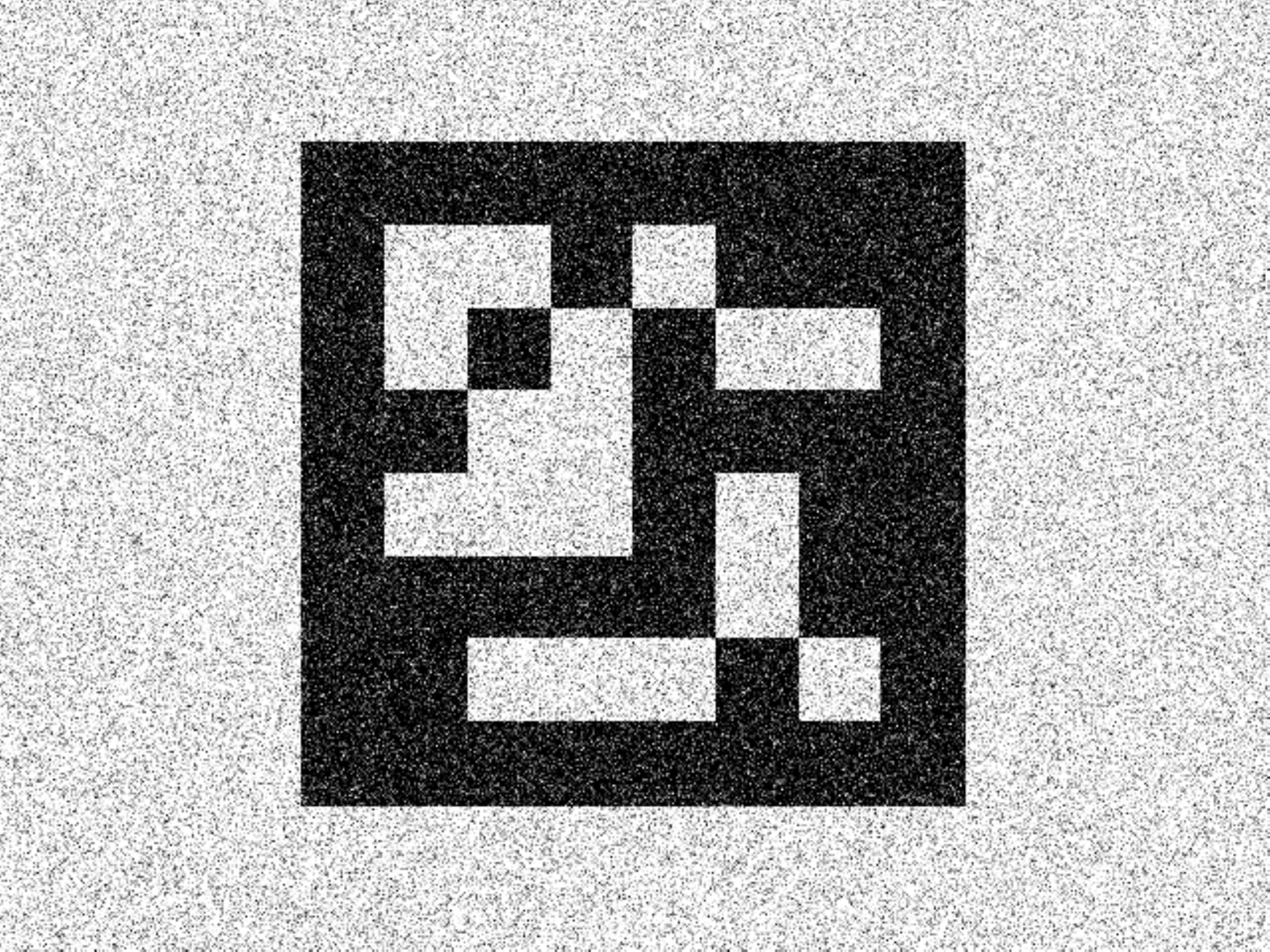}
  \centerline{(b)}
\end{minipage}
\begin{minipage}[b]{0.45\linewidth} 
  \centering
  \includegraphics[width=\linewidth]{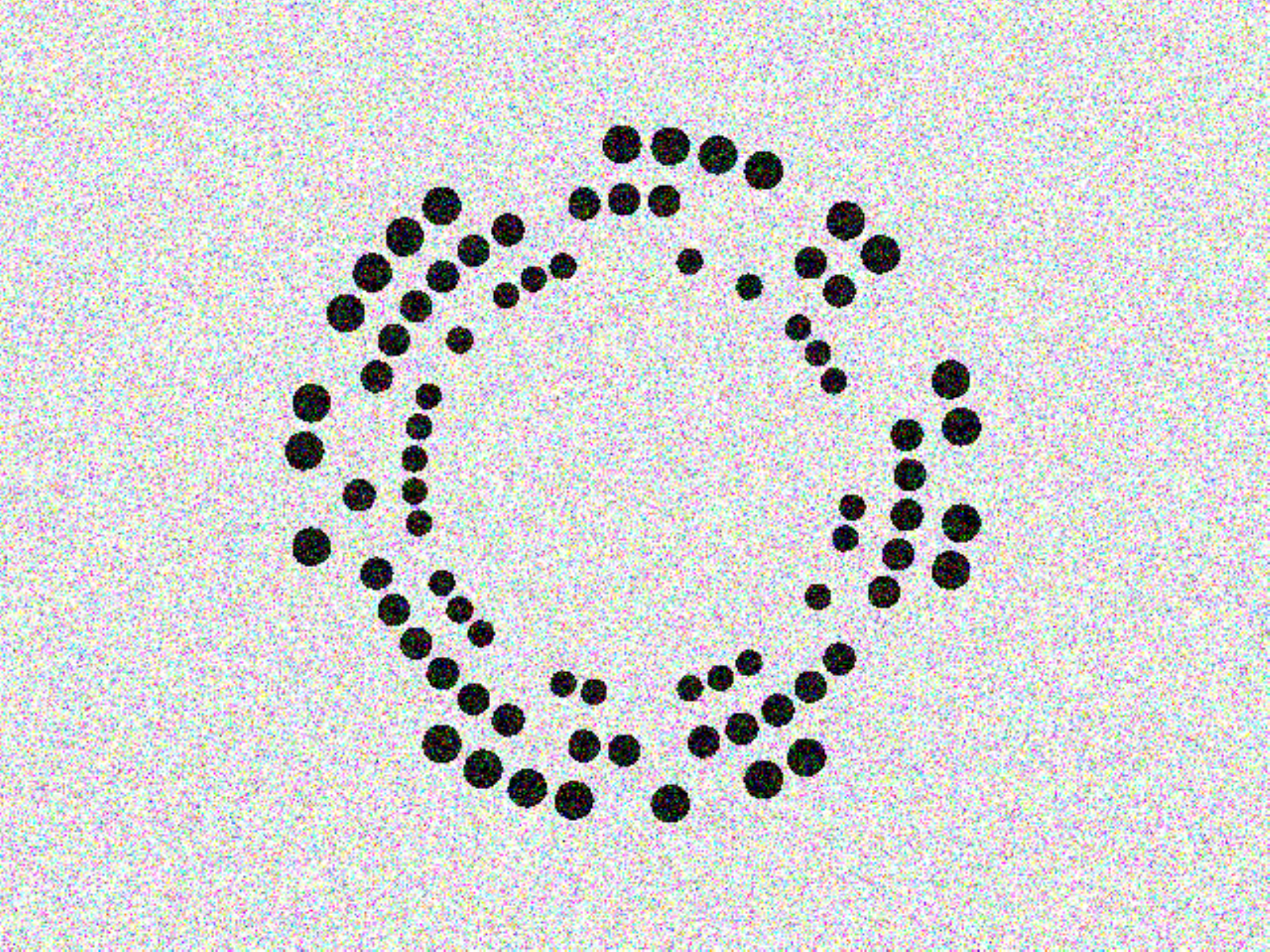}
  \centerline{(c)}
\end{minipage}
\begin{minipage}[b]{0.45\linewidth} 
  \centering
  \includegraphics[width=\linewidth]{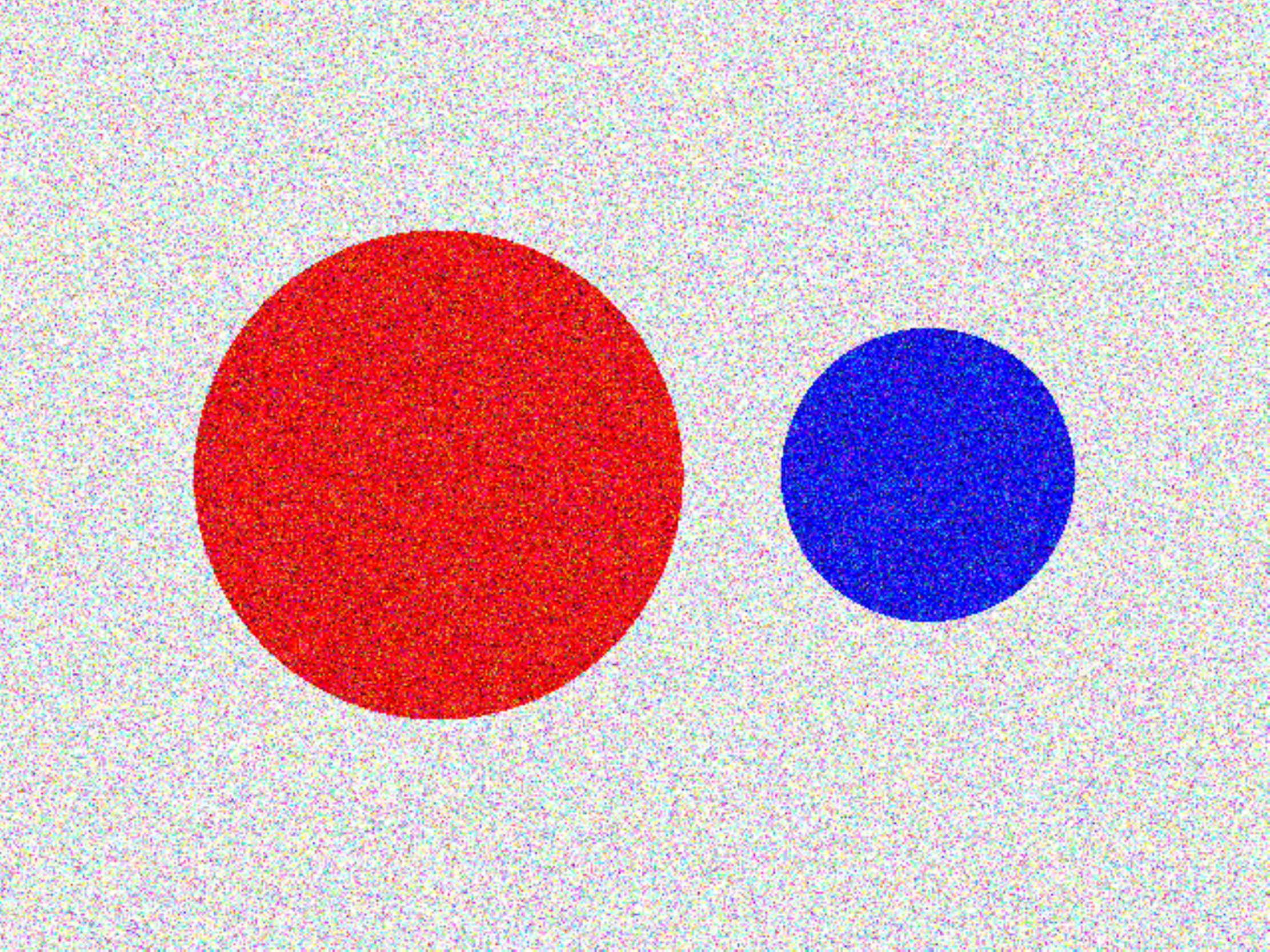}
  \centerline{(d)}
\end{minipage}
\begin{minipage}[b]{1.0\linewidth}
  \centering
  \includegraphics[width=\linewidth]{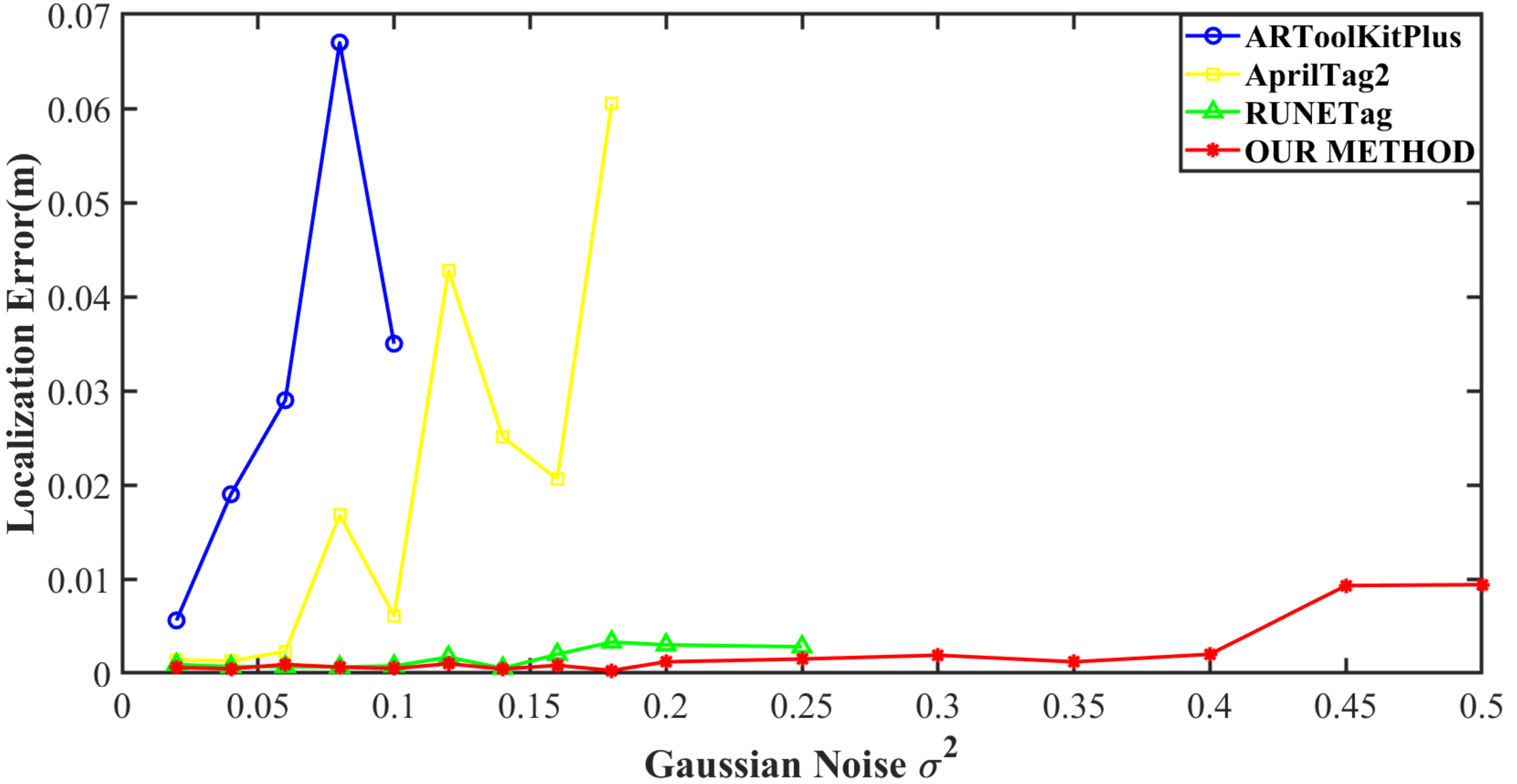}
  \centerline{(e)}
\end{minipage}
\caption{Evaluation VS. different Gaussian noise levels: (a) ARToolkitPlus with Gaussian noise level 0.1; (b) AprilTag2 with Gaussian noise level 0.1; (c) RUNETag with Gaussian noise level 0.1 ; (d) The marker of our method with Gaussian noise level 0.1; (e) Localization accuracies via Gaussian noise levels.}
\label{fig:long}
\label{fig:onecol}
\end{figure}
\begin{figure}[t]
\centering
\begin{minipage}[b]{0.45\linewidth}  
  \centering
  \includegraphics[width=\linewidth]{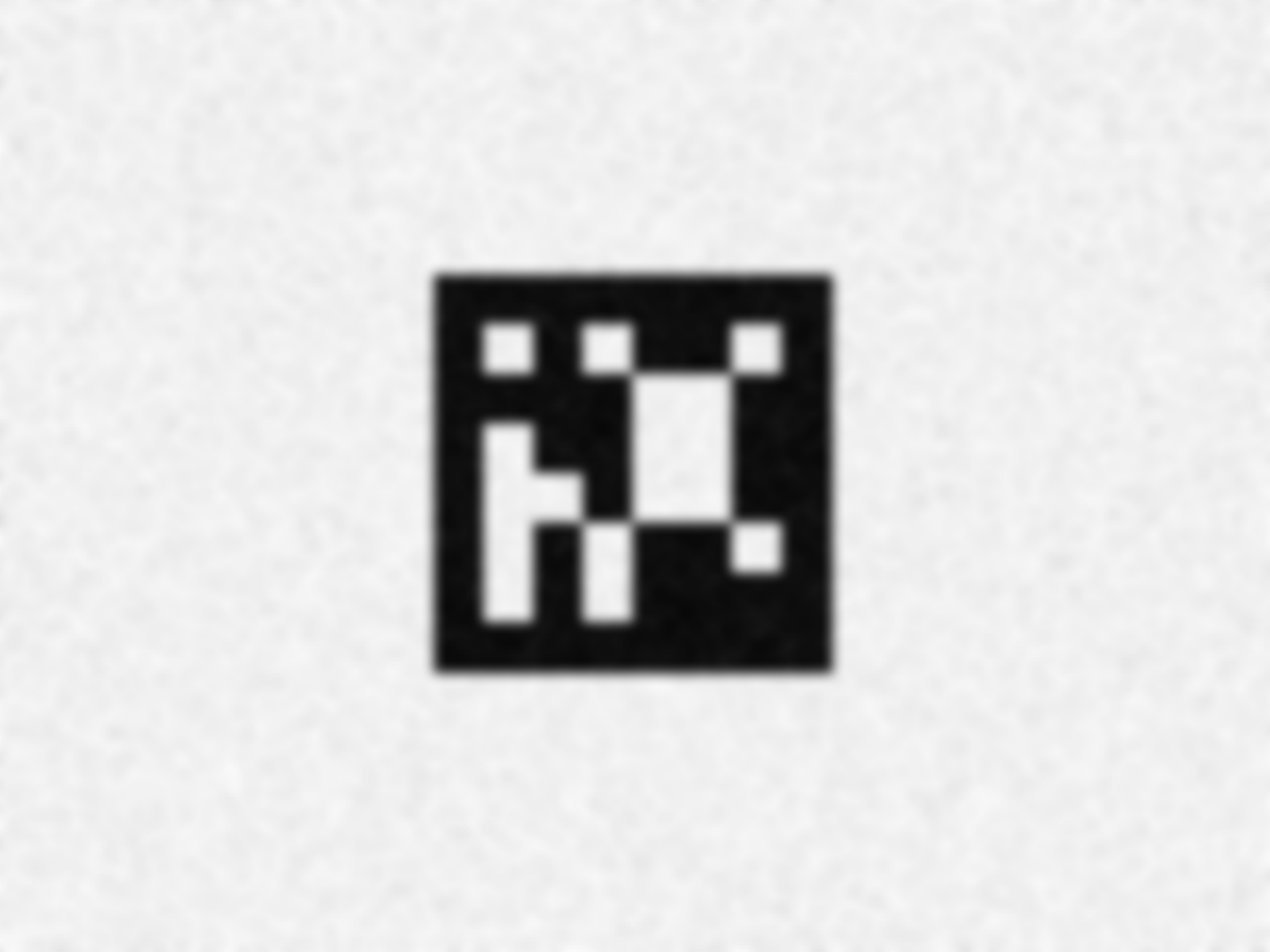}
  \centerline{(a)}
\end{minipage}
\begin{minipage}[b]{0.45\linewidth}
  \centering
  \includegraphics[width=\linewidth]{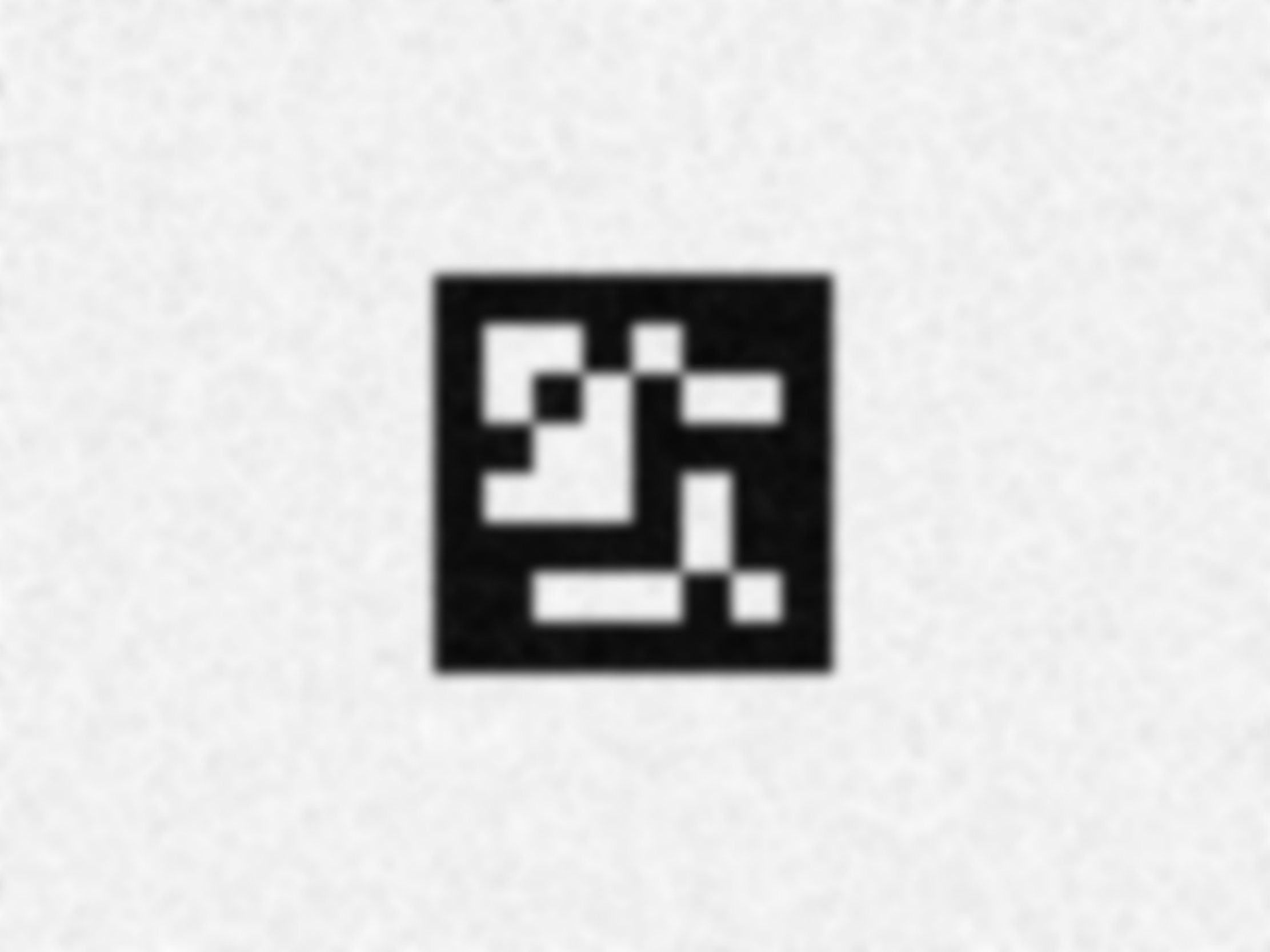}
  \centerline{(b)}
\end{minipage}
\begin{minipage}[b]{0.45\linewidth}
  \centering
  \includegraphics[width=\linewidth]{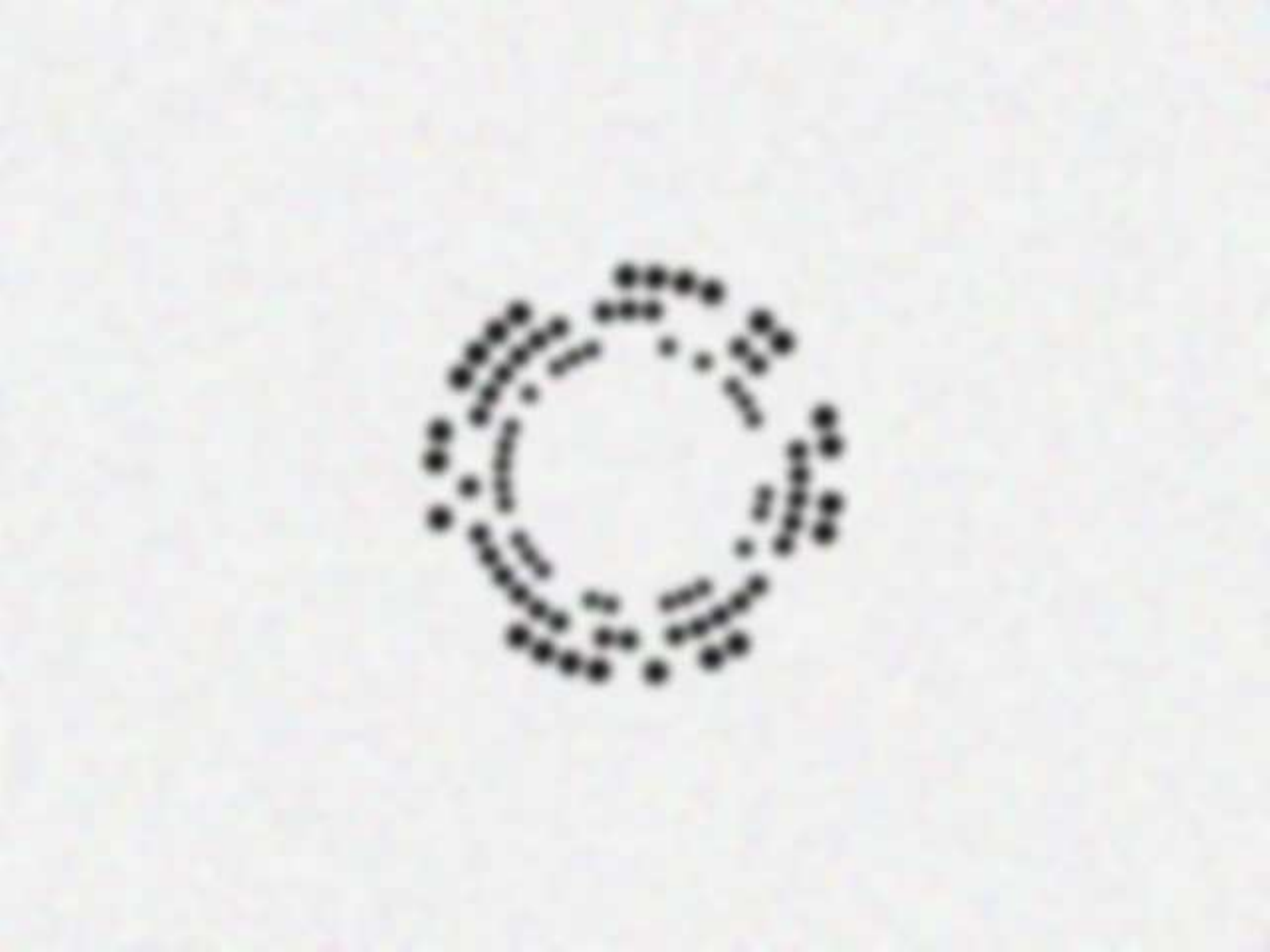}
  \centerline{(c)}
\end{minipage}
\begin{minipage}[b]{0.45\linewidth}
  \centering
  \includegraphics[width=\linewidth]{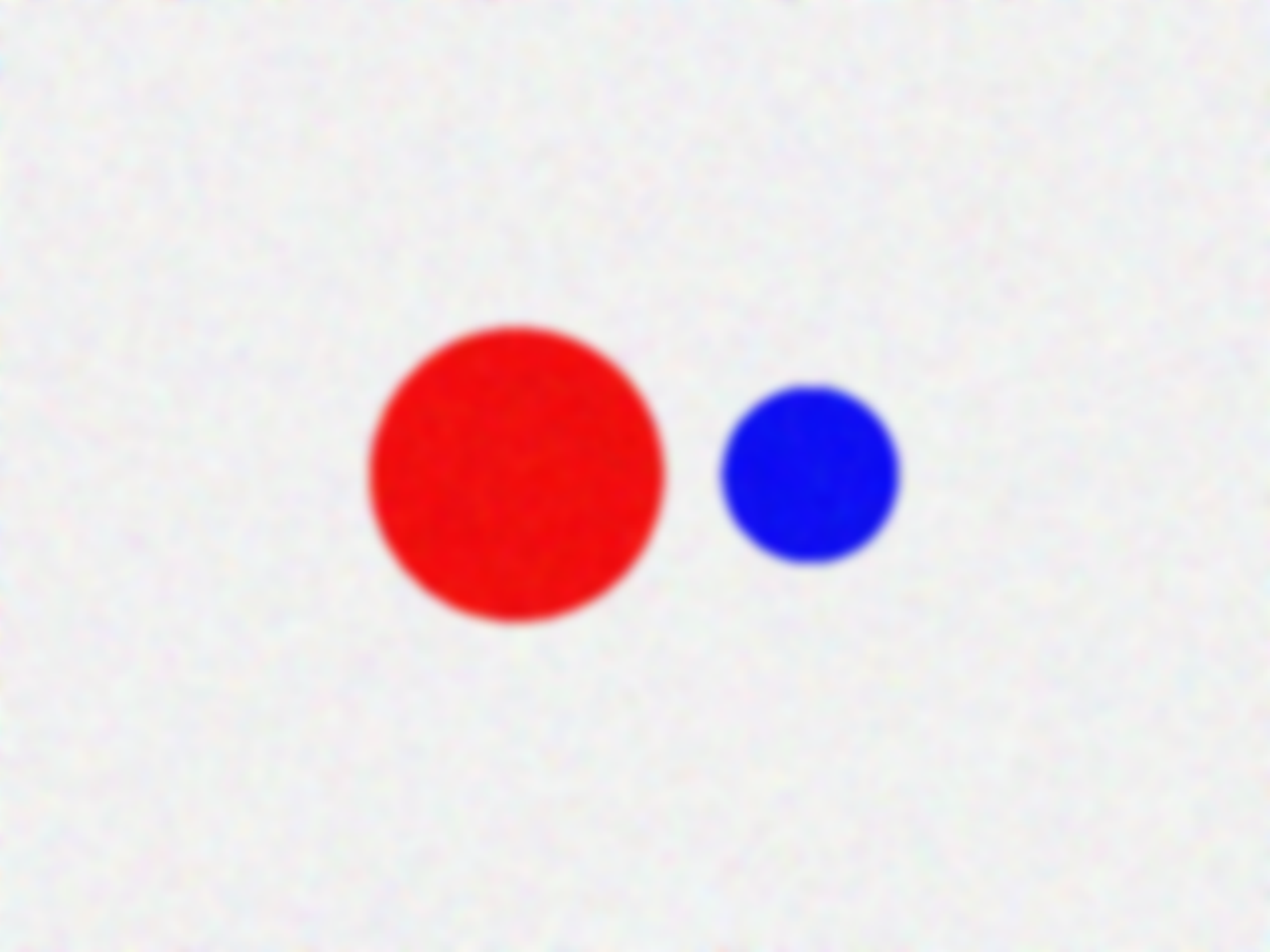}
  \centerline{(d)}
\end{minipage}
\begin{minipage}[b]{1.0\linewidth}
  \centering
  \includegraphics[width=\linewidth]{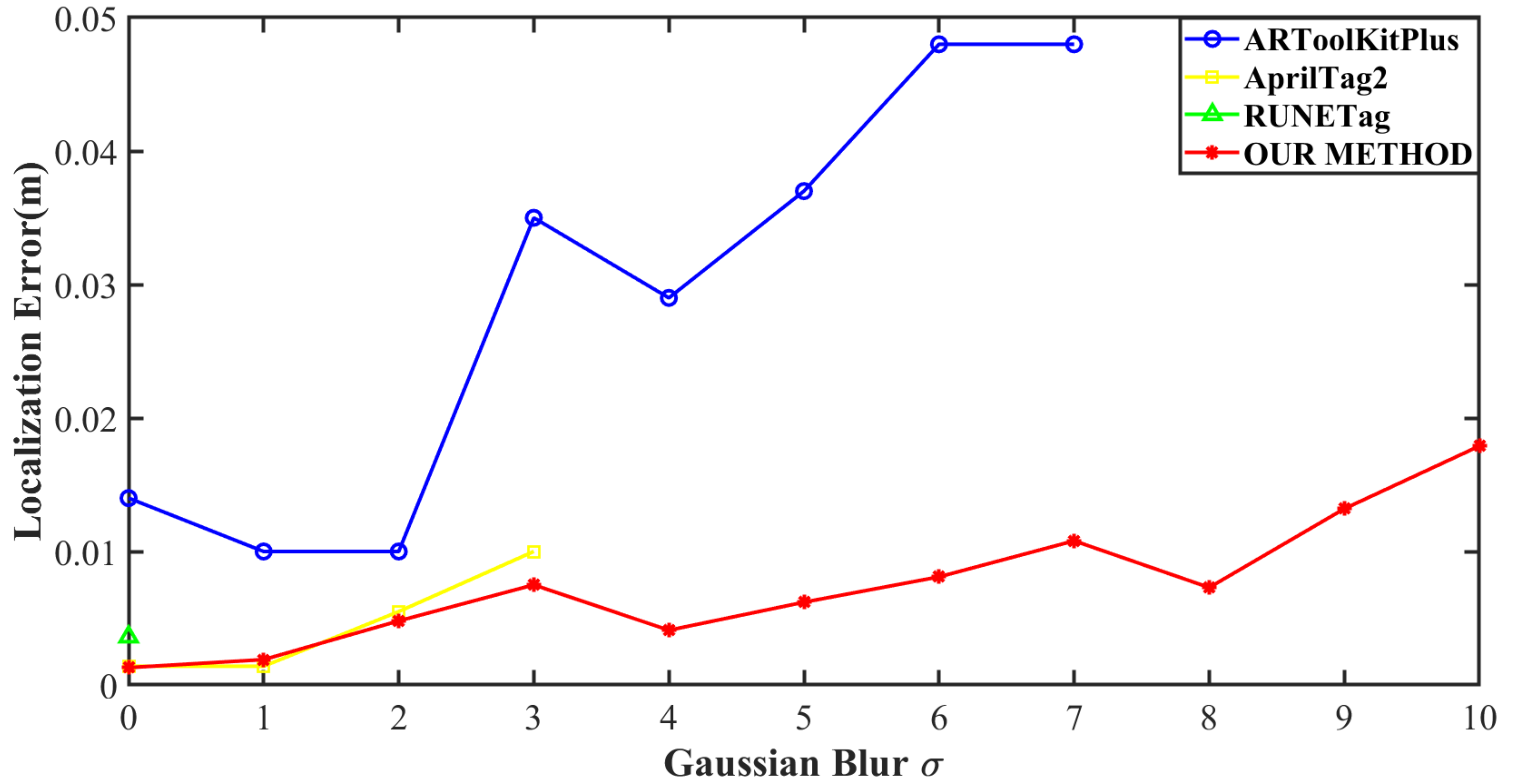}
  \centerline{(e)}
\end{minipage}
\caption{ Evaluation VS. different Gaussian blur levels: (a) ARToolkitPlus with Gaussian blur level ${\sigma} = 3$ and with Gaussian noise level 0.02; (b) AprilTag2 with the same Gaussian blur and noise as (a); (c) RUNETag with the same Gaussian blur and noise; (d) The marker of our method with the same Gaussian blur and noise; (e) Localization accuracies via Gaussian blur levels.}
\label{fig:long}
\label{fig:onecol}
\end{figure}
\begin{figure}[t]
\centering
\begin{minipage}[b]{0.45\linewidth}   
  \centering
  \includegraphics[width=\linewidth]{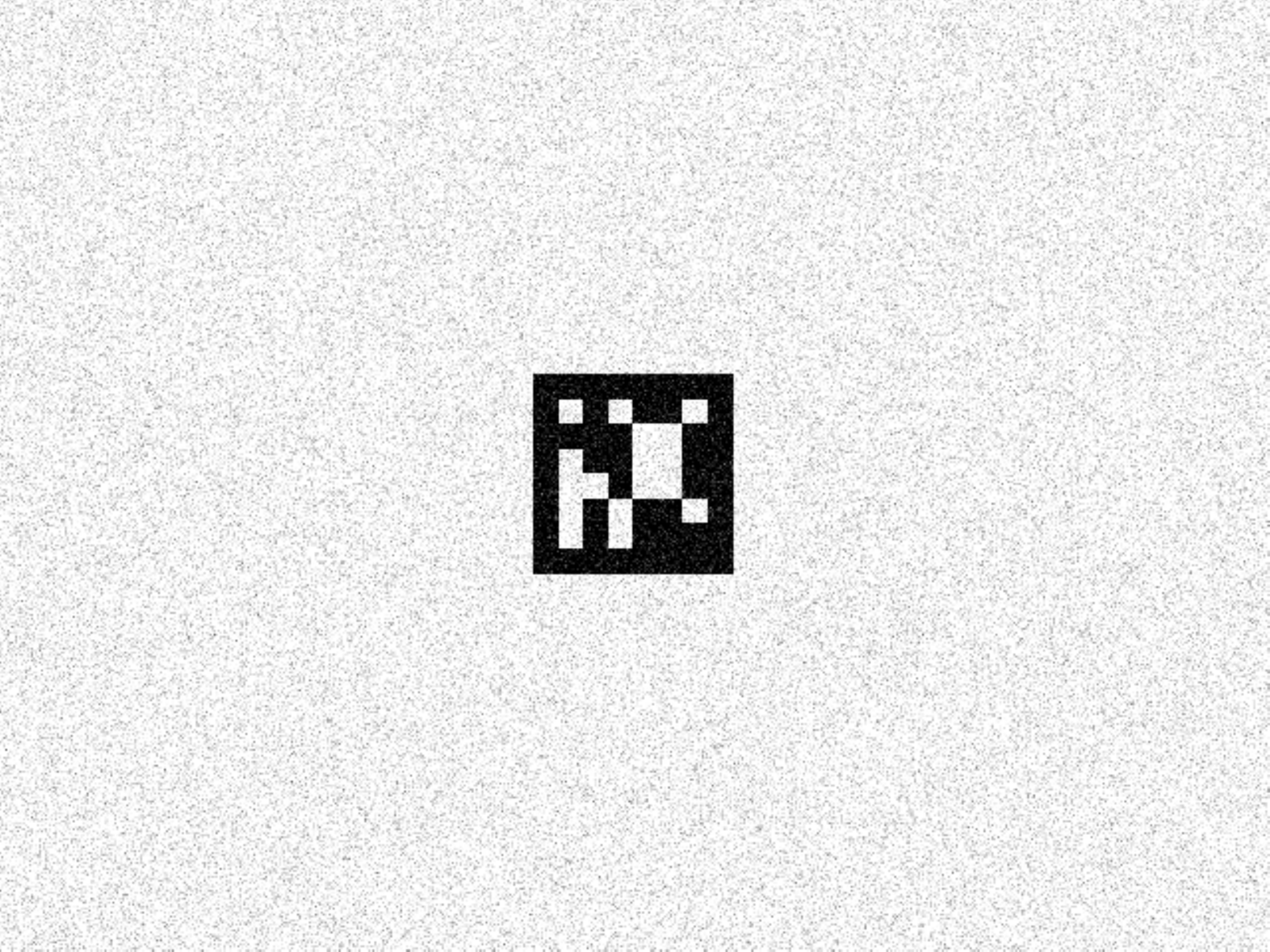}
  \centerline{(a)}
\end{minipage}
\begin{minipage}[b]{0.45\linewidth}
  \centering
  \includegraphics[width=\linewidth]{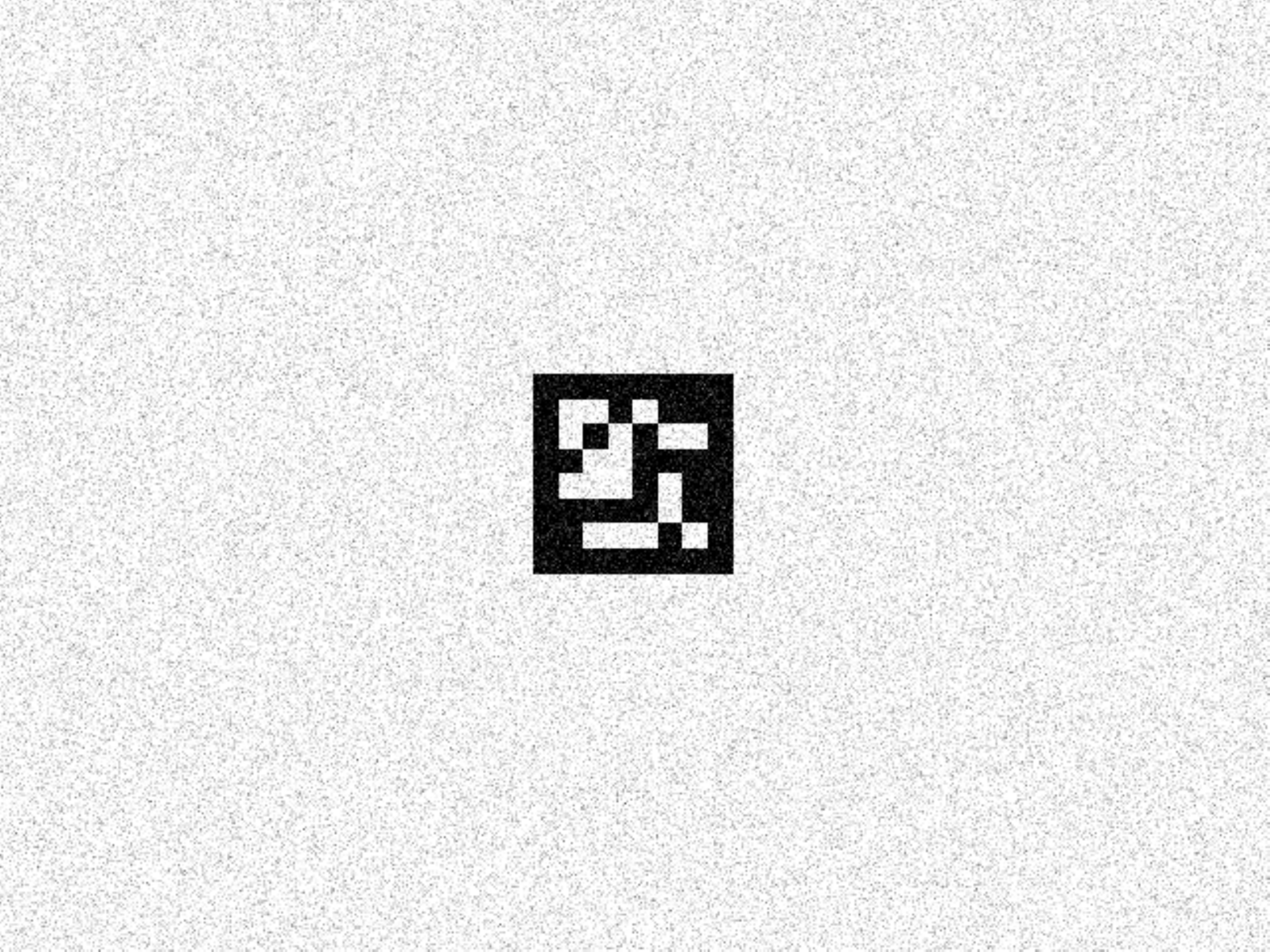}
  \centerline{(b)}
\end{minipage}
\begin{minipage}[b]{0.45\linewidth}
  \centering
  \includegraphics[width=\linewidth]{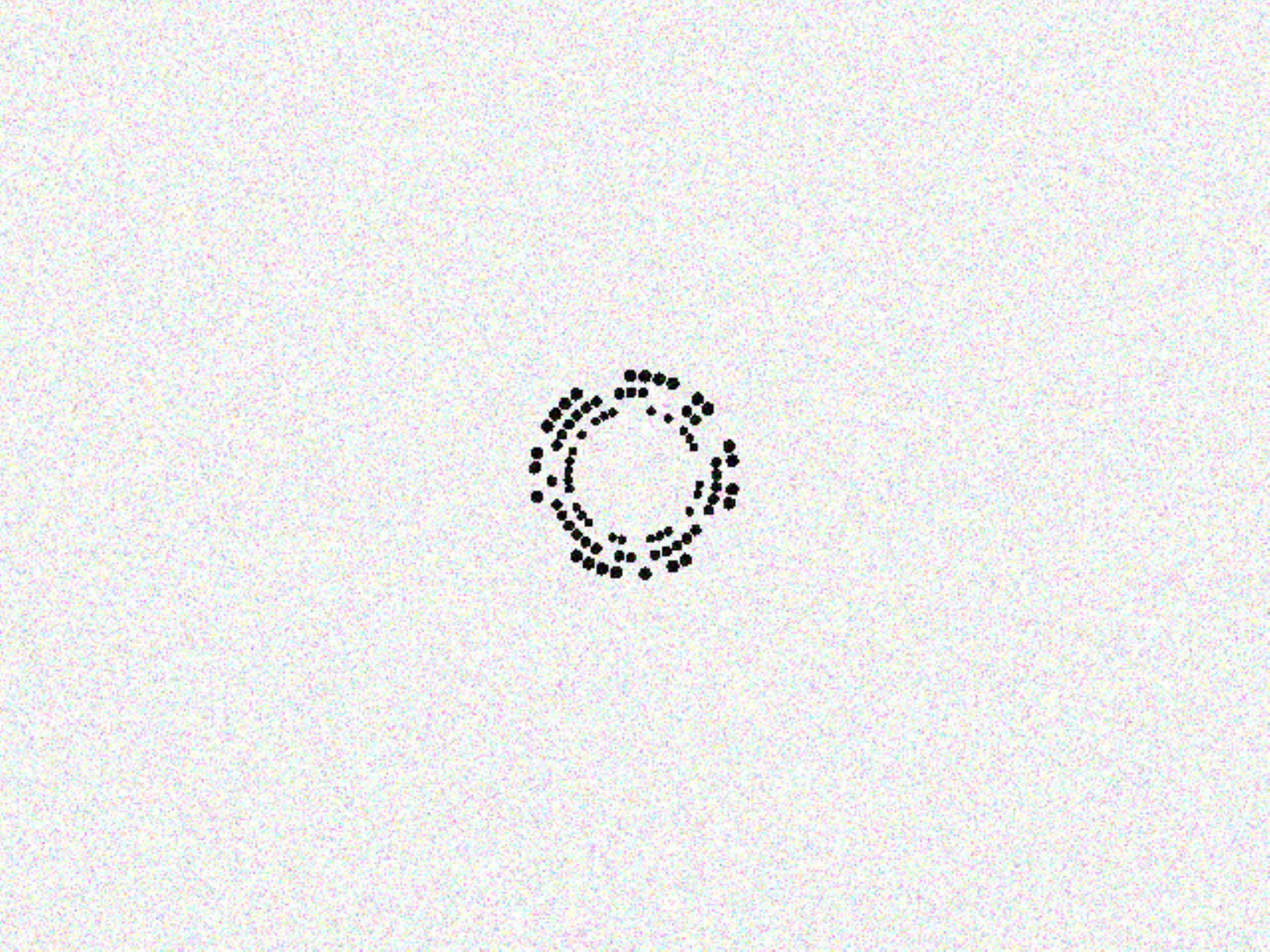}
  \centerline{(c)}
\end{minipage}
\begin{minipage}[b]{0.45\linewidth}
  \centering
  \includegraphics[width=\linewidth]{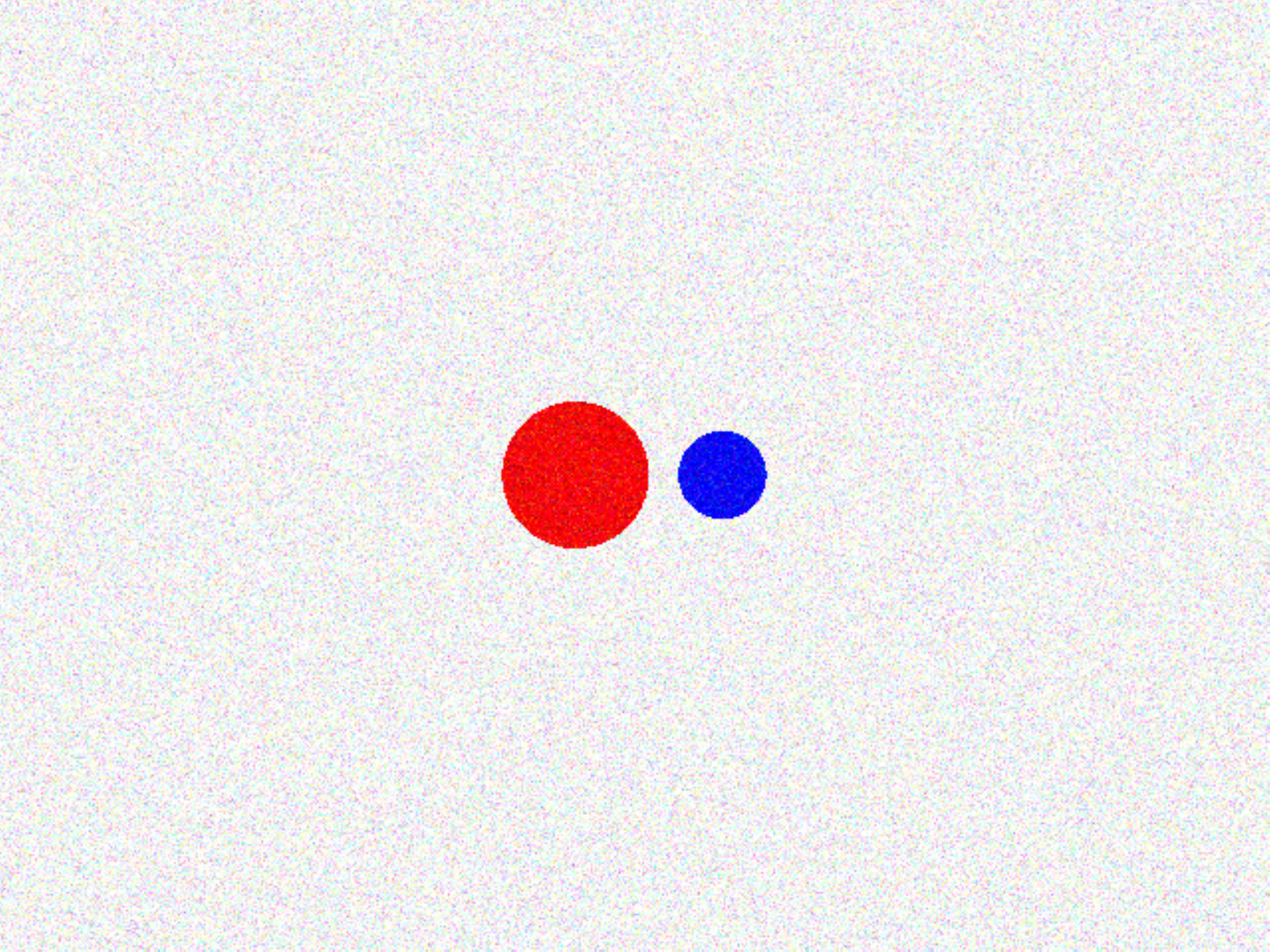}
  \centerline{(d)}
\end{minipage}
\begin{minipage}[b]{1.0\linewidth}
  \centering
  \includegraphics[width=\linewidth]{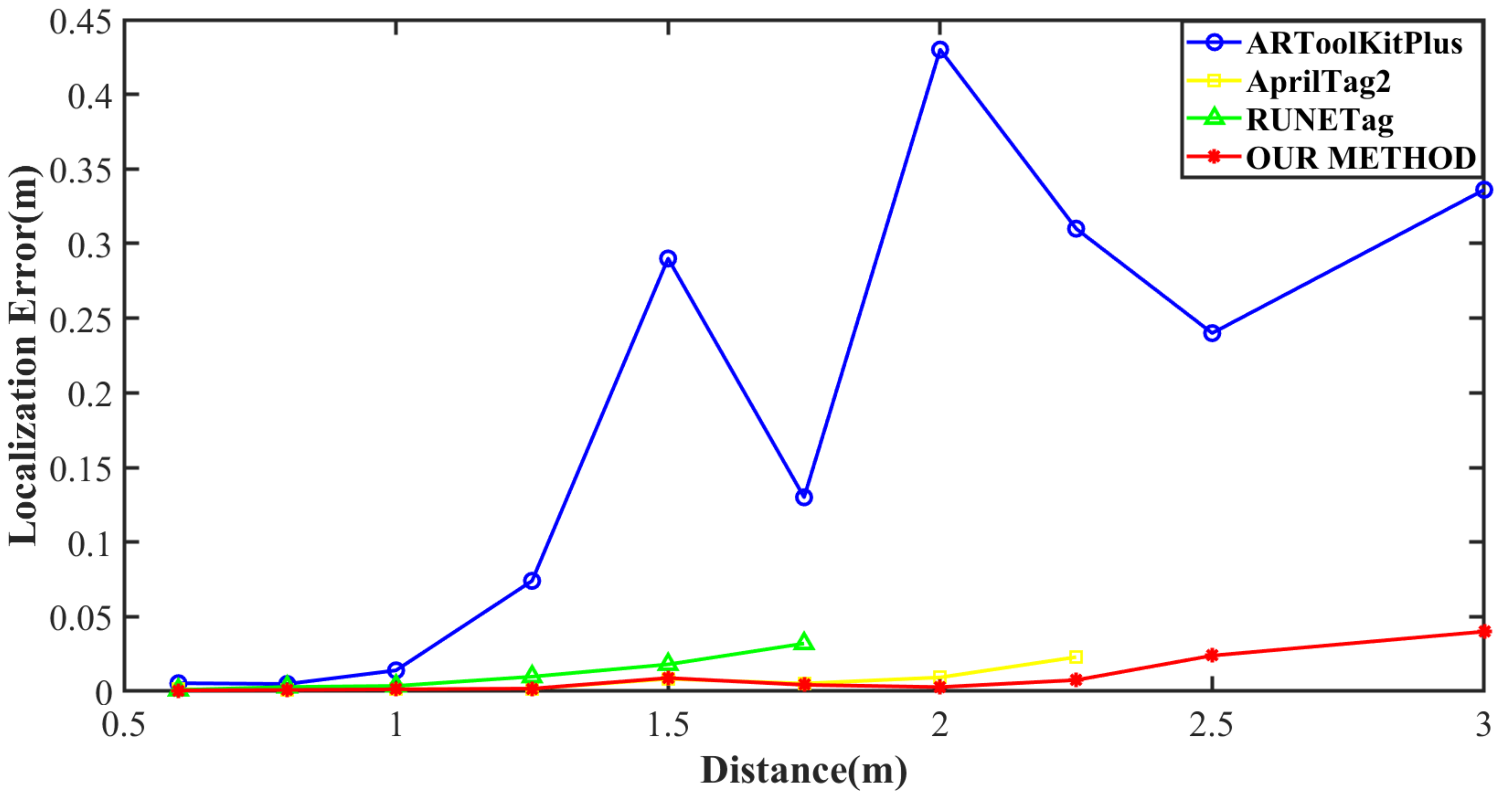}
  \centerline{(e)}
\end{minipage}
\caption{Evaluation VS. different distances from camera to the marker: (a) ARToolkitPlus with Gaussian noise level 0.02 and with the distance 2.0m from camera to the marker; (b) AprilTag2 with the same condition as (a). (c) RUNETag with the same condition; (d) The marker of our method with the same condition; (e) Localization accuracies via different distances from camera to the marker.}
\label{fig:long}
\label{fig:onecol}
\end{figure}

From Section 4 and Section 5, we can see that we do not use PnP and RANSAC techniques, usually used by most camera pose tracking methods from markers. Our method is not only accurate and robust but also fast at the same time. Experiments in the next section including comparisons with ARToolkitPlus \cite{wagner2010real}, AprilTag2 \cite{wang2016apriltag} and RUNETag \cite{bergamasco2016accurate} prove the remarks.

\section{Experiments}
We compare our method with ARToolkitPlus \cite{wagner2010real}, AprilTag2 \cite{wang2016apriltag} and RUNETag \cite{bergamasco2016accurate} on image frames with Gaussian noise, Gaussian blur, and different distances from camera to marker. ARToolkitPlus and AprilTag2 are considered as a de-facto standard markers for augmented reality applications. RUNETag is an accurate and robust artificial marker based on cyclic codes, published in T-PAMI in 2016. In order to get ground truths, we simulate images of 640x480 pixels from markers of ARToolkitPlus, AprilTag2, RUNETag and our method. To be fair, as it is shown in Figure 5, the same image area of different markers is set. What's more, the same camera pose of them is set. The results show that our method outperforms ARToolkitPlus, AprilTag2 and RUNETag on Gaussian noise, Gaussian blur and different distances from the camera to markers.

Furthermore, we perform experiments on real images and make augmented reality for evaluating robustness and speed. When lights in the scene change and images get blurry, our method can still run stably. Moreover, our method can run at about 100 FPS on a consumer computer. The speed is much faster than RUNETag, which runs at less than 5 FPS on the same computer.

We also make comparison with  ORBSLAM \cite{mur2015orb, mur2017orb}, showing that the camera trajectory of our method is more smooth than the camera trajectory of ORBSLAM. The high robustness and accuracies of our method make that the marker of our method can be put into a natural environment and the results of camera localization can be used as an evaluation criterion for general SLAM systems under natural environments without ground truth.
\begin{figure*}[t]
\centering
\begin{minipage}[b]{0.24\linewidth}
  \centering
  \includegraphics[width=\linewidth]{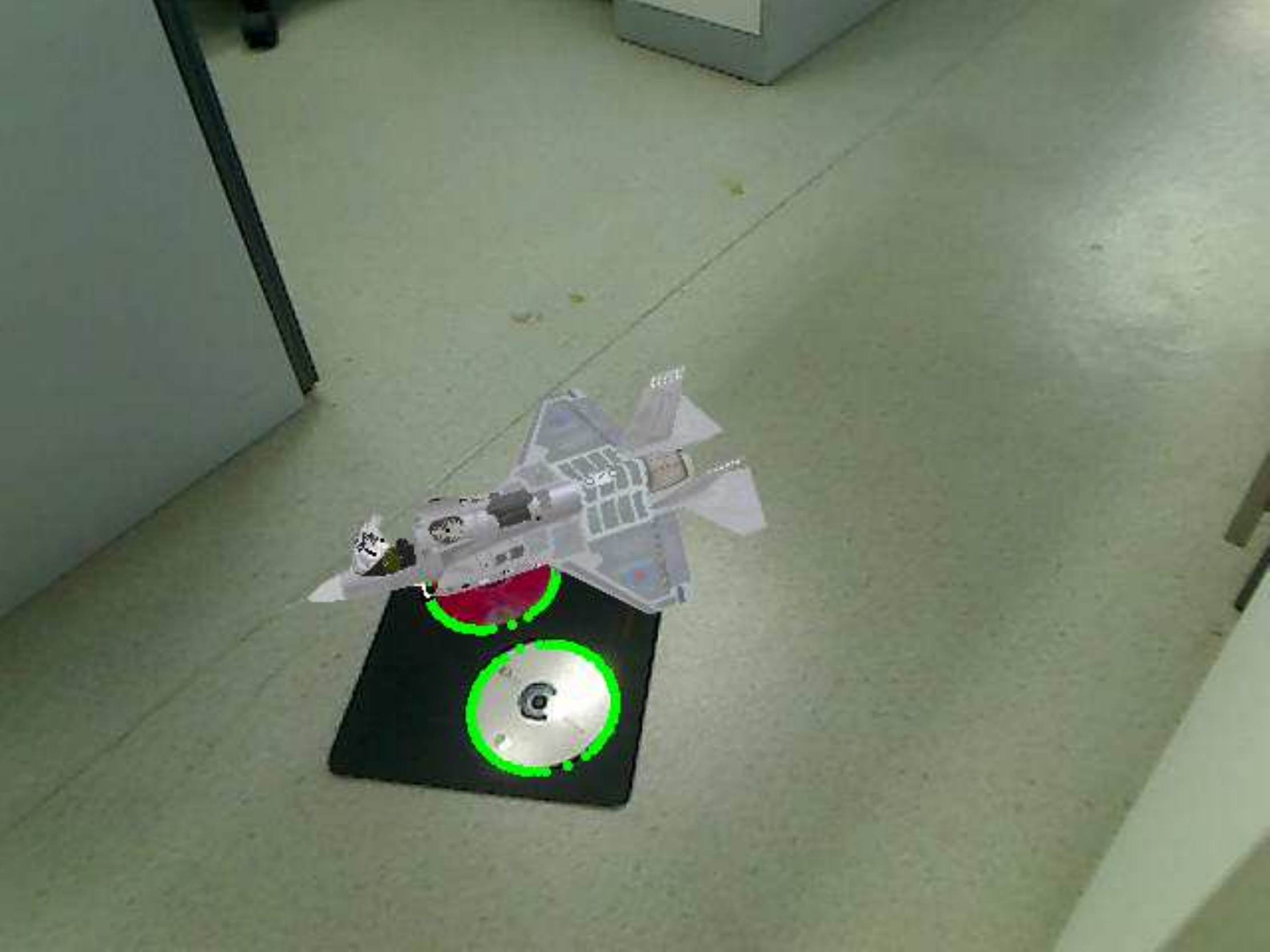}
  \centerline{(a)}
\end{minipage}
\begin{minipage}[b]{0.24\linewidth}
  \centering
  \includegraphics[width=\linewidth]{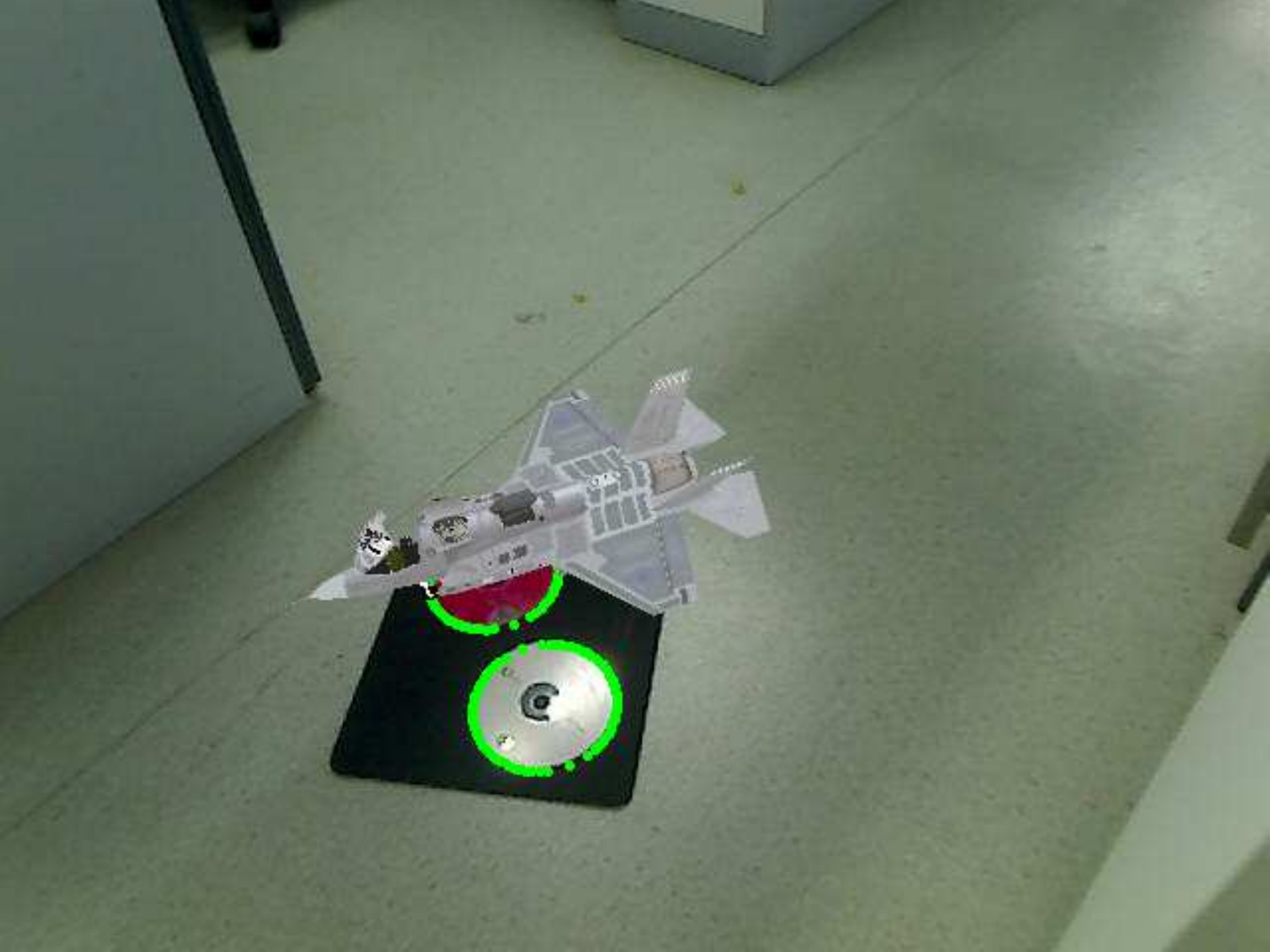}
  \centerline{(b)}
\end{minipage}
\begin{minipage}[b]{0.24\linewidth}
  \centering
  \includegraphics[width=\linewidth]{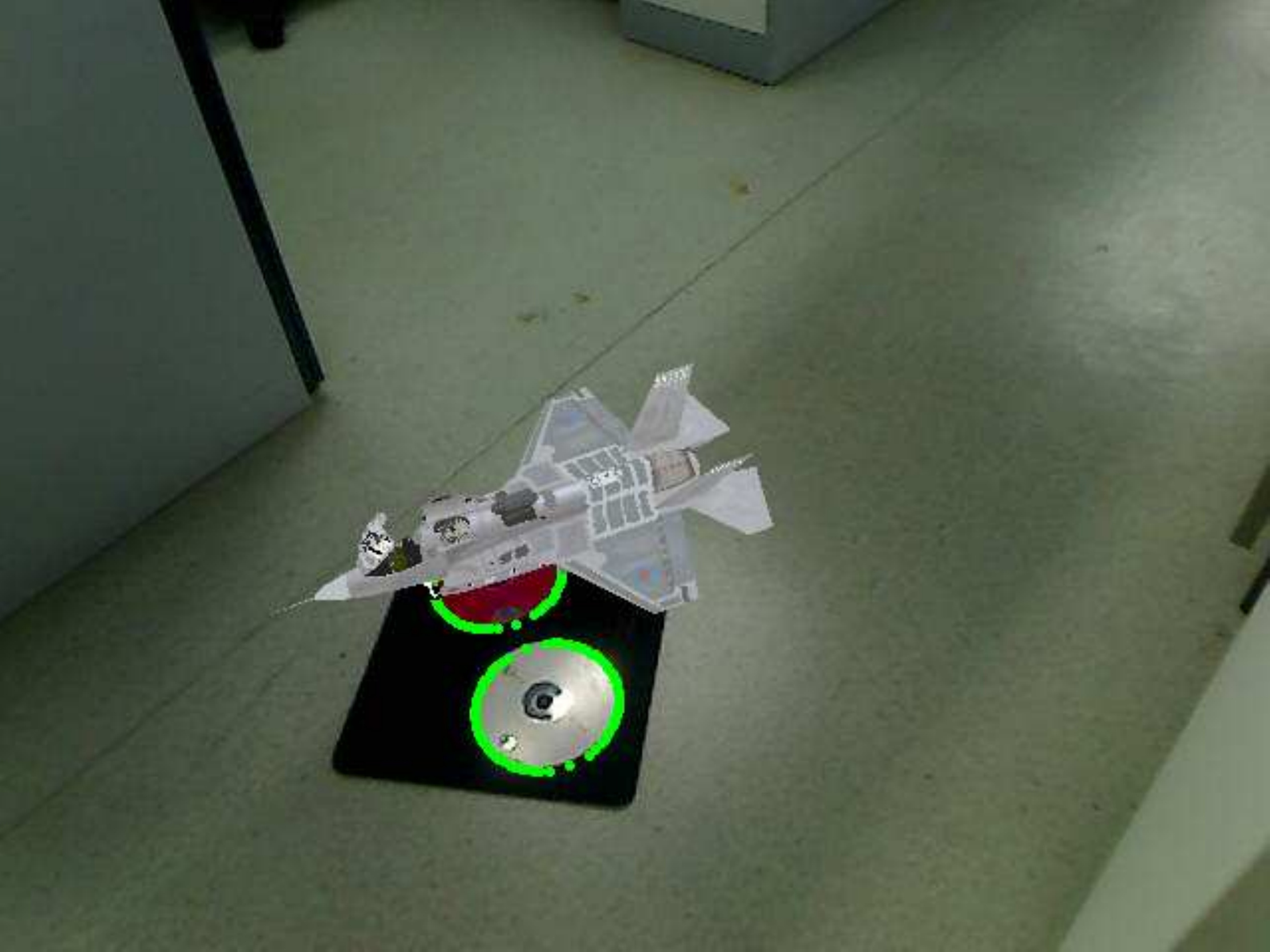}
  \centerline{(c)}
\end{minipage}
\begin{minipage}[b]{0.24\linewidth}
  \centering
  \includegraphics[width=\linewidth]{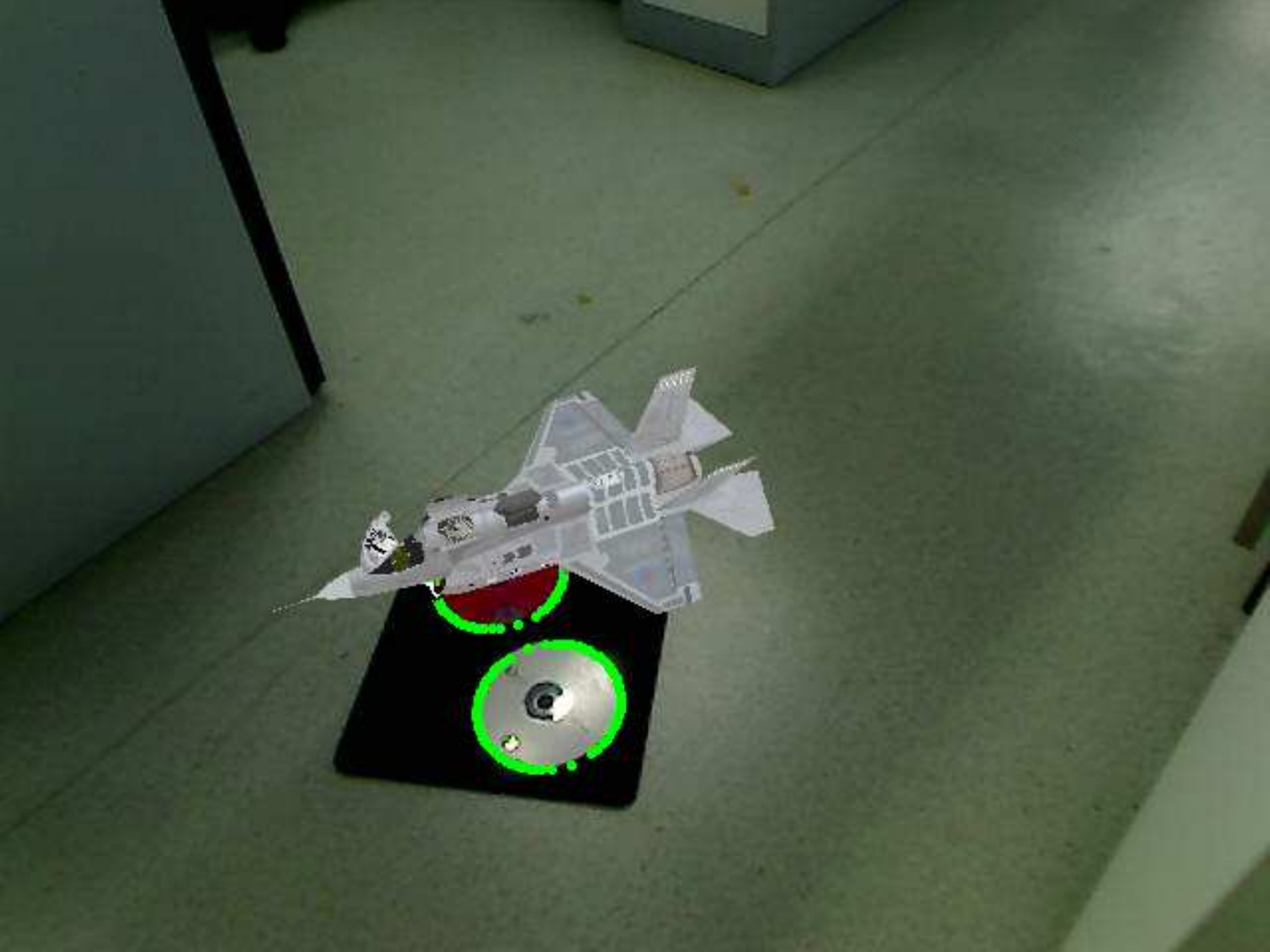}
  \centerline{(d)}
\end{minipage}
\caption{Augmented reality with different illuminations, where the airplane is virtual. From (a) to (d), illuminations are changed. The scene is brighter in (a) and (b) and darker in (c) and (d).}
\label{fig:long}
\label{fig:onecol}
\end{figure*}
\begin{figure*}[t]
\centering
\begin{minipage}[b]{0.24\linewidth}
  \centering
  \includegraphics[width=\linewidth]{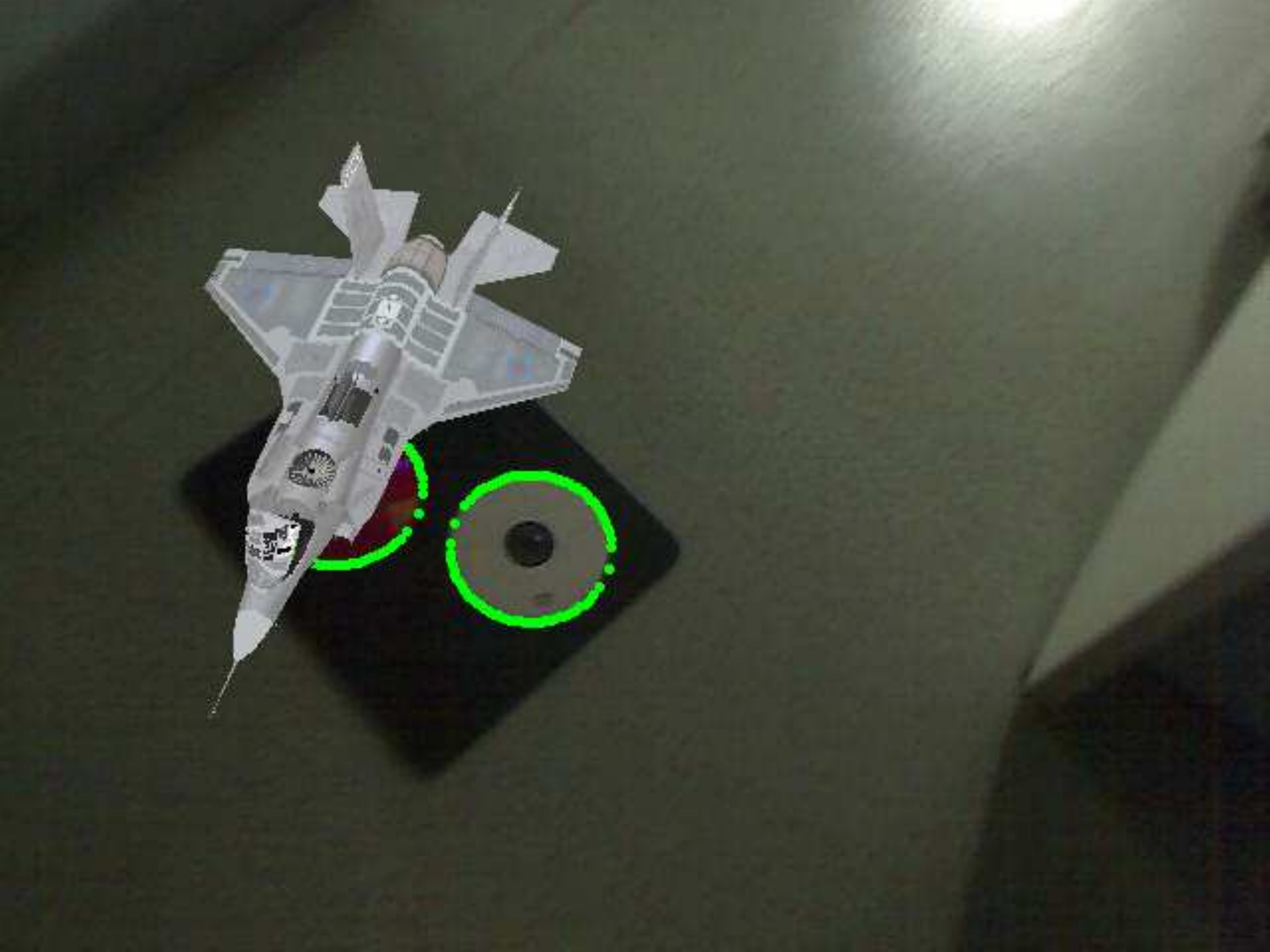}
  \centerline{(a)}
\end{minipage}
\begin{minipage}[b]{0.24\linewidth}
  \centering
  \includegraphics[width=\linewidth]{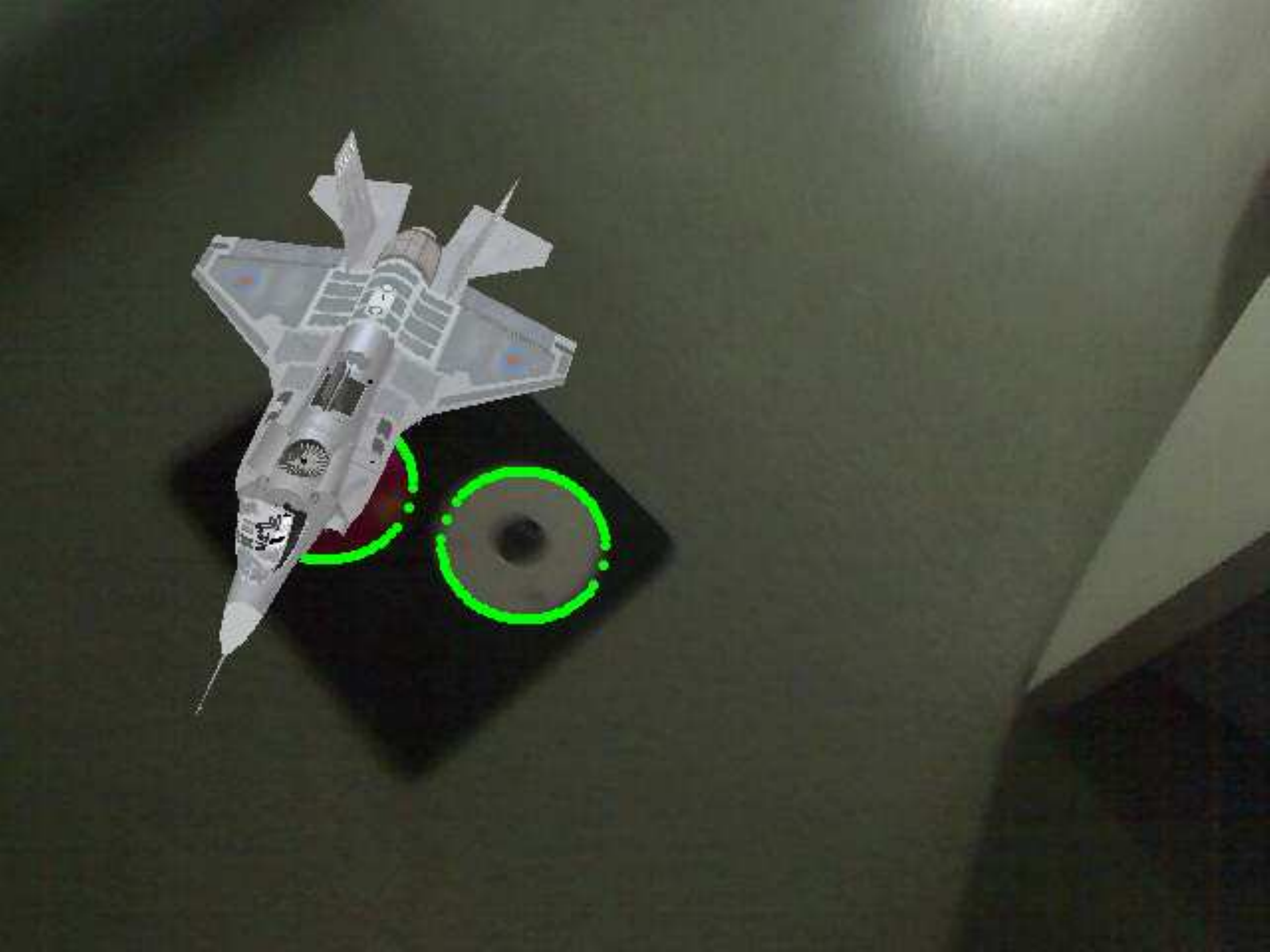}
  \centerline{(b)}
\end{minipage}
\begin{minipage}[b]{0.24\linewidth}
  \centering
  \includegraphics[width=\linewidth]{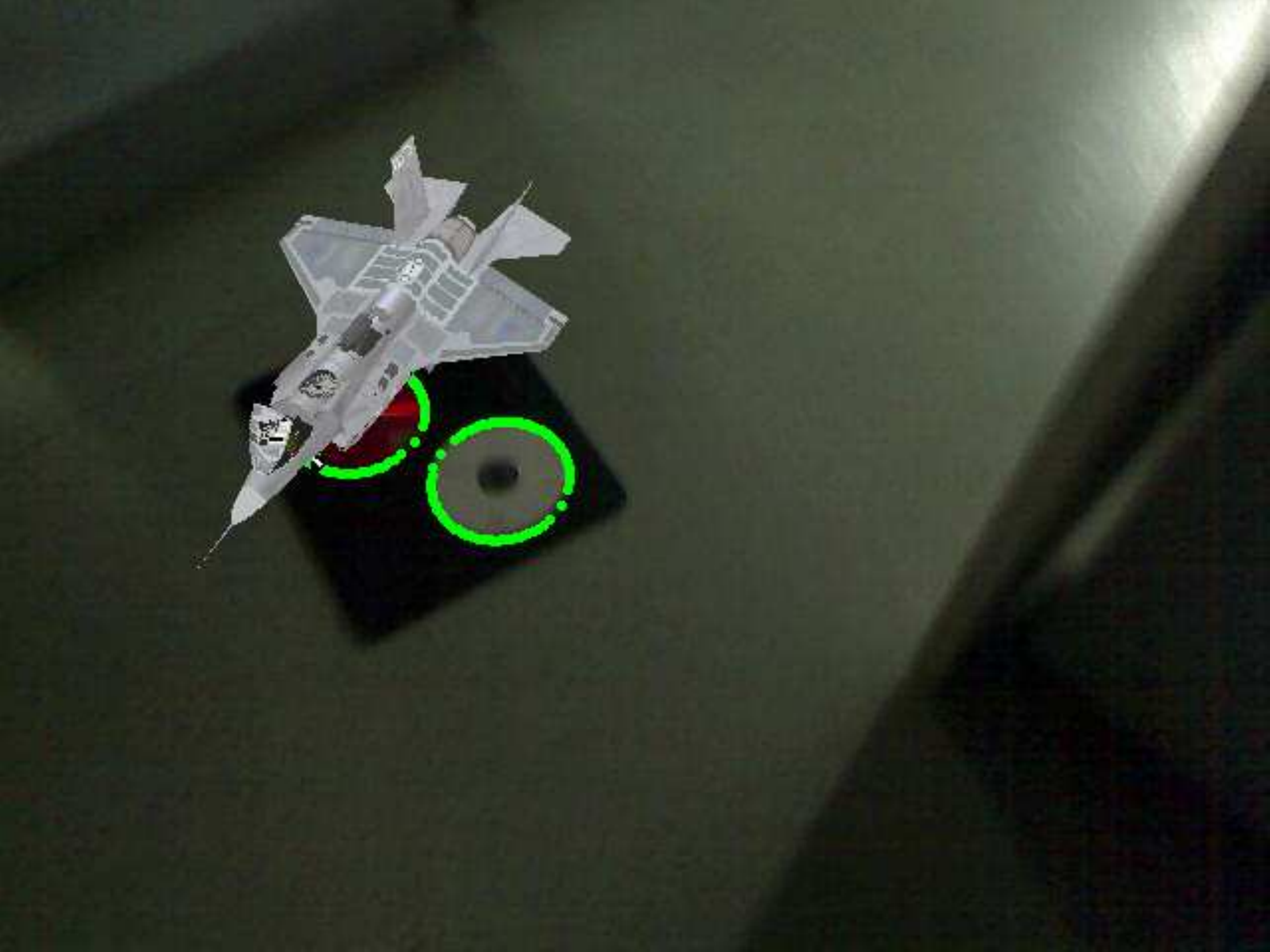}
  \centerline{(c)}
\end{minipage}
\begin{minipage}[b]{0.24\linewidth}
  \centering
  \includegraphics[width=\linewidth]{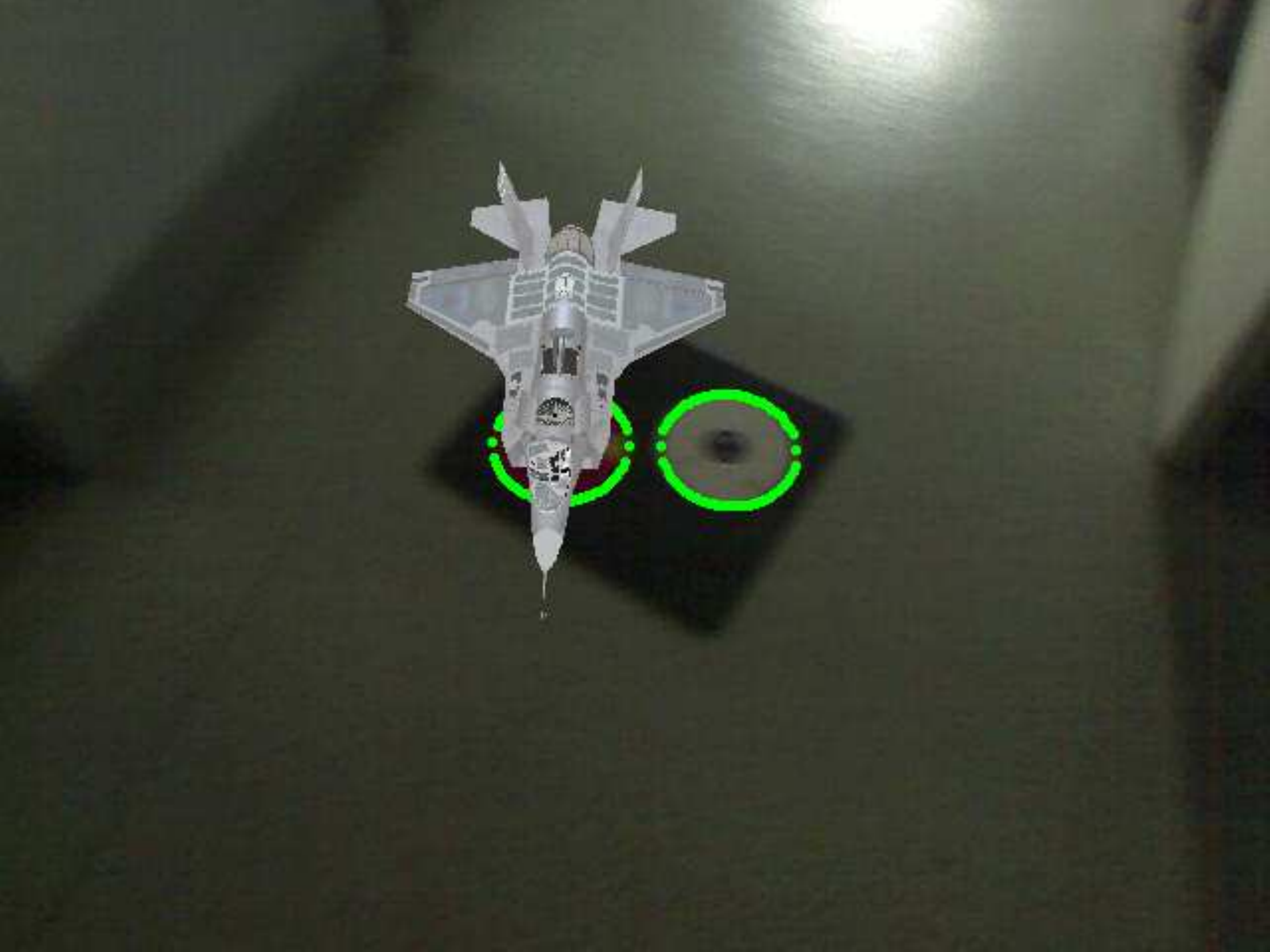}
  \centerline{(d)}
\end{minipage}
\caption{Augmented reality with different blur levels, where the airplane is virtual. From (a) to (d), images are more and more blur in a dark scene. The camera pose can still be recovered and the virtual airplane is placed stably.}
\label{fig:long}
\label{fig:onecol}
\end{figure*}

\subsection{Evaluation with Comparisons}
Figure 6, Figure 7 and Figure 8 show the localization accuracies of ARToolkitPlus, AprilTag2, RUNETag and our method vs. Gaussian noise, Gaussian blur and distances from camera to marker respectively. The localization error used is Euclidean distance between the computed camera optical center and its ground truth.

At first, we evaluate with Gaussian noise. (a), (b), (c) and (d) of Figure 6 show images of different markers with Gaussian noise level ${\sigma}^2 = 0.1$ and with the distance 0.6m from camera to marker. (e) of Figure 6 shows the localization accuracies of ARToolkitPlus, AprilTag2, RUNETag and our method vs. different Gaussian noise levels. We can see that ARToolkitPlus and AprilTag2 are not stable. When variance ${\sigma}^2$ of Gaussian noise is more than 0.1, ARToolkitPlus fails. When variance ${\sigma}^2$ of Gaussian noise is more than 0.18, AprilTag2 fails. The main reason is that the markers can not be detected and recognized. RUNETag is more robust to noise than ARToolkitPlus and AprilTag2. With increasing variance ${\sigma}^2$ of Gaussian noise, although there are wrong ellipses detected, it can be implemented successfully and obtain good localization accuracy by RUNETag. This mainly accounts for the fact that RANSAC is used to remove outliers and PnP is used to refine camera pose. Our method has the similar accuracies with RUNETag when variance ${\sigma}^2$ of Gaussian noise is small. With variance ${\sigma}^2$ of Gaussian noise increasing, the accuracy of our method outperforms RUNETag slightly. When variance ${\sigma}^2$ of Gaussian noise is more than 0.25, RUNETag fails and the principal cause is that there are few right ellipses detected. However, our method can still be successful and stable all along.

Secondly, we add Gaussian blur to images with standard deviations varying from 0 to 10. (a), (b), (c) and (d) of Figure 7 show examples of images with ${\sigma} = 3$ of Gaussian blur and with the distance 1.0m from camera to the marker. (e) of Figure 7 shows the localization accuracies of ARToolkitPlus, AprilTag2, RUNETag and our method vs. different Gaussian blur levels. We can see that RUNETag can be successful only with zero blur. ARToolkitPlus fails when standard deviation ${\sigma}$ of Gaussian blur is more than 7. AprilTag2 fails when standard deviation ${\sigma}$ of Gaussian blur is more than 3. It is obvious that our method obtains the highest accuracies with the most stability.

Finally, we vary the distance from camera to marker and then localize the camera. (a), (b), (c) and (d) of Figure 8 show examples of images with the distance 2.0m and with Gaussian noise level ${\sigma}^2 = 0.02$. (e) of Figure 8 shows the localization accuracies of ARToolkitPlus, AprilTag2, RUNETag and our method via different distances from camera to marker. When the distance from camera to the marker is small, AprilTag2, RUNETag and our method have similar localization accuracies. With increasing the distance, localization of our method is still stable with high accuracies. However, ARToolkitPlus is unstable and has large errors. AprilTag fails when the distance is more than 2.25m. RUNETag fails when the distance is more than 1.75m.
\begin{figure*}[t]
\centering
\begin{minipage}[b]{0.24\linewidth}
  \centering
  \includegraphics[width=\linewidth]{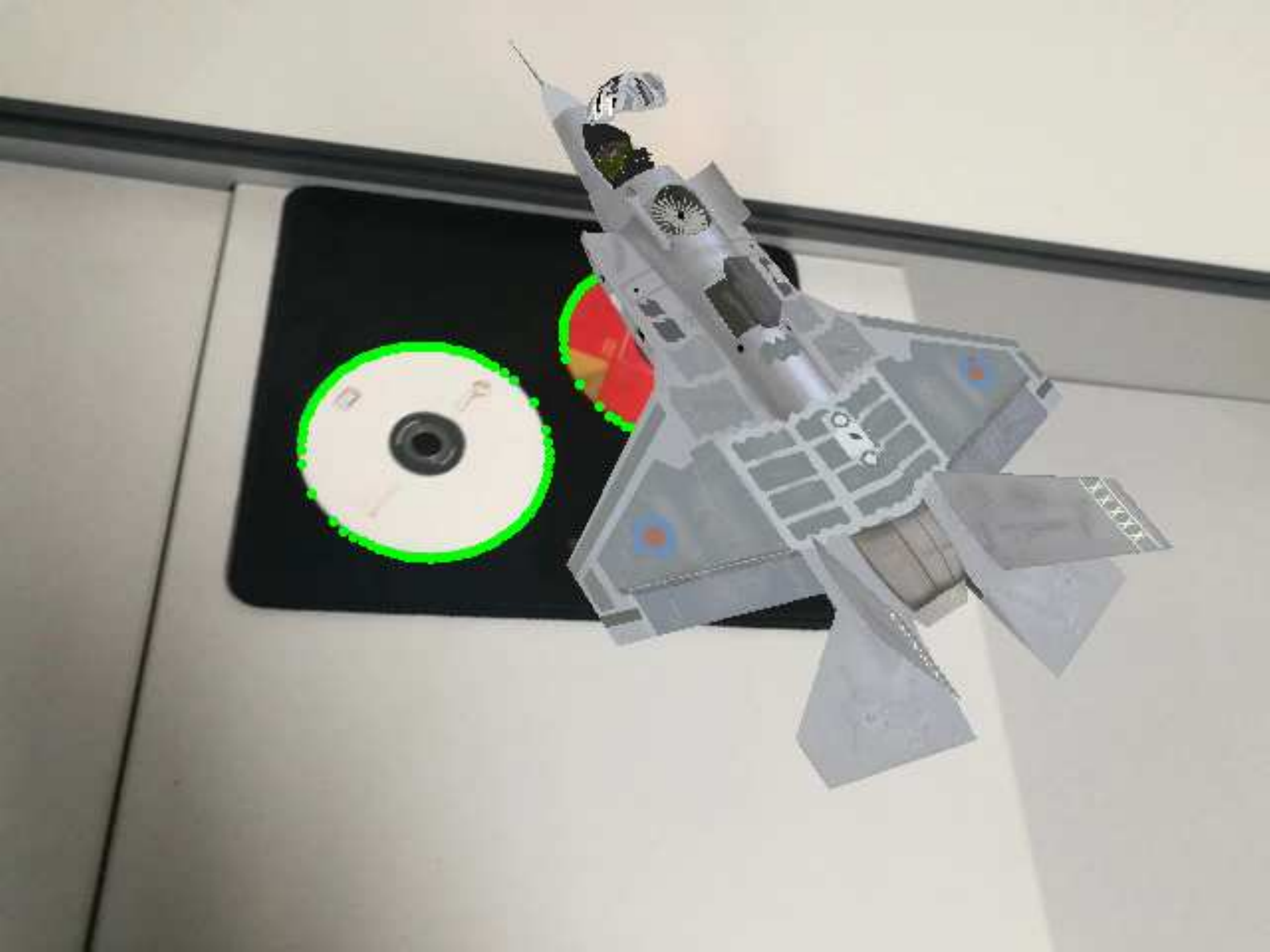}
  \centerline{(a)}
\end{minipage}
\begin{minipage}[b]{0.24\linewidth}
  \centering
  \includegraphics[width=\linewidth]{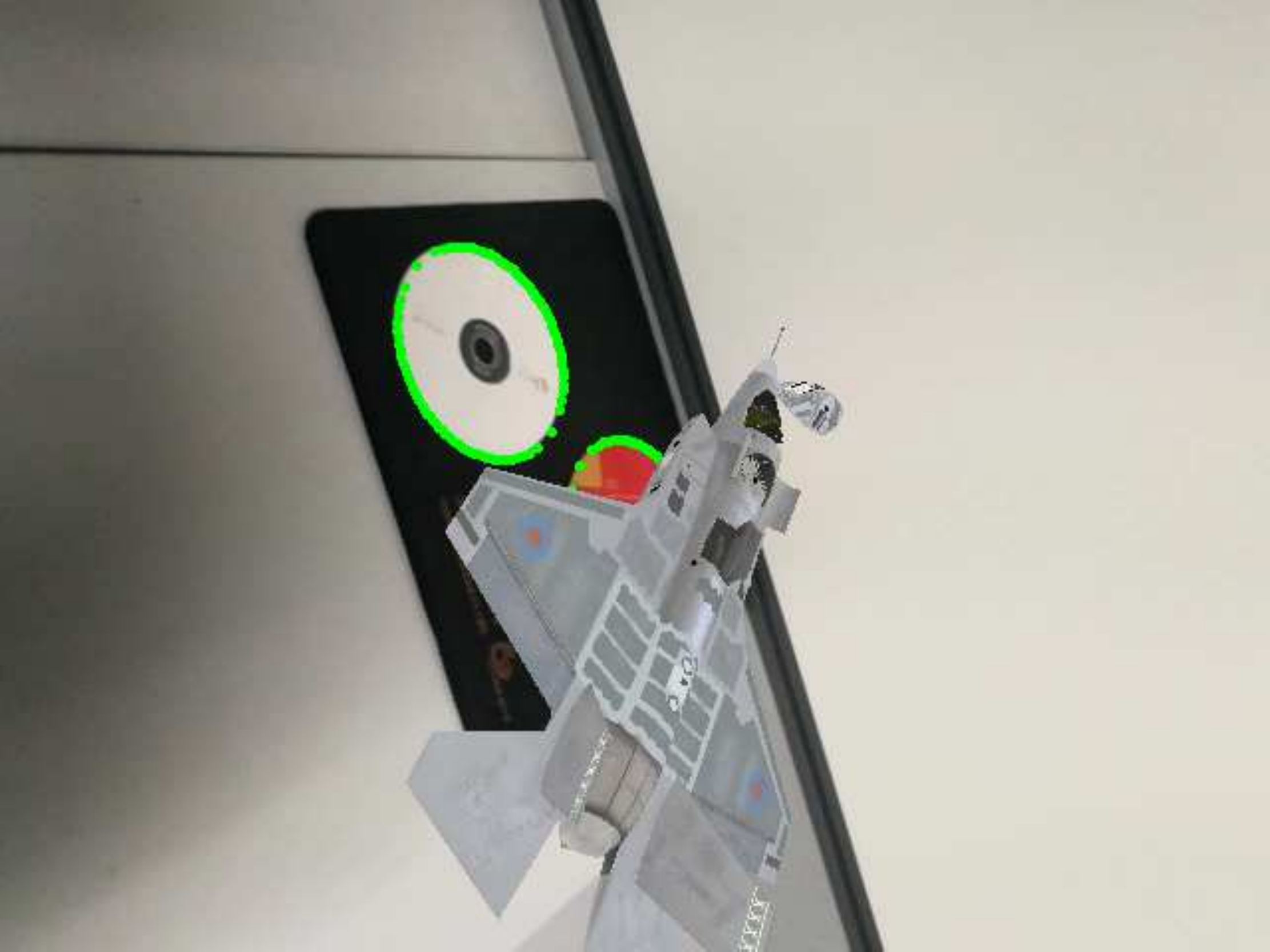}
  \centerline{(b)}
\end{minipage}
\begin{minipage}[b]{0.24\linewidth}
  \centering
  \includegraphics[width=\linewidth]{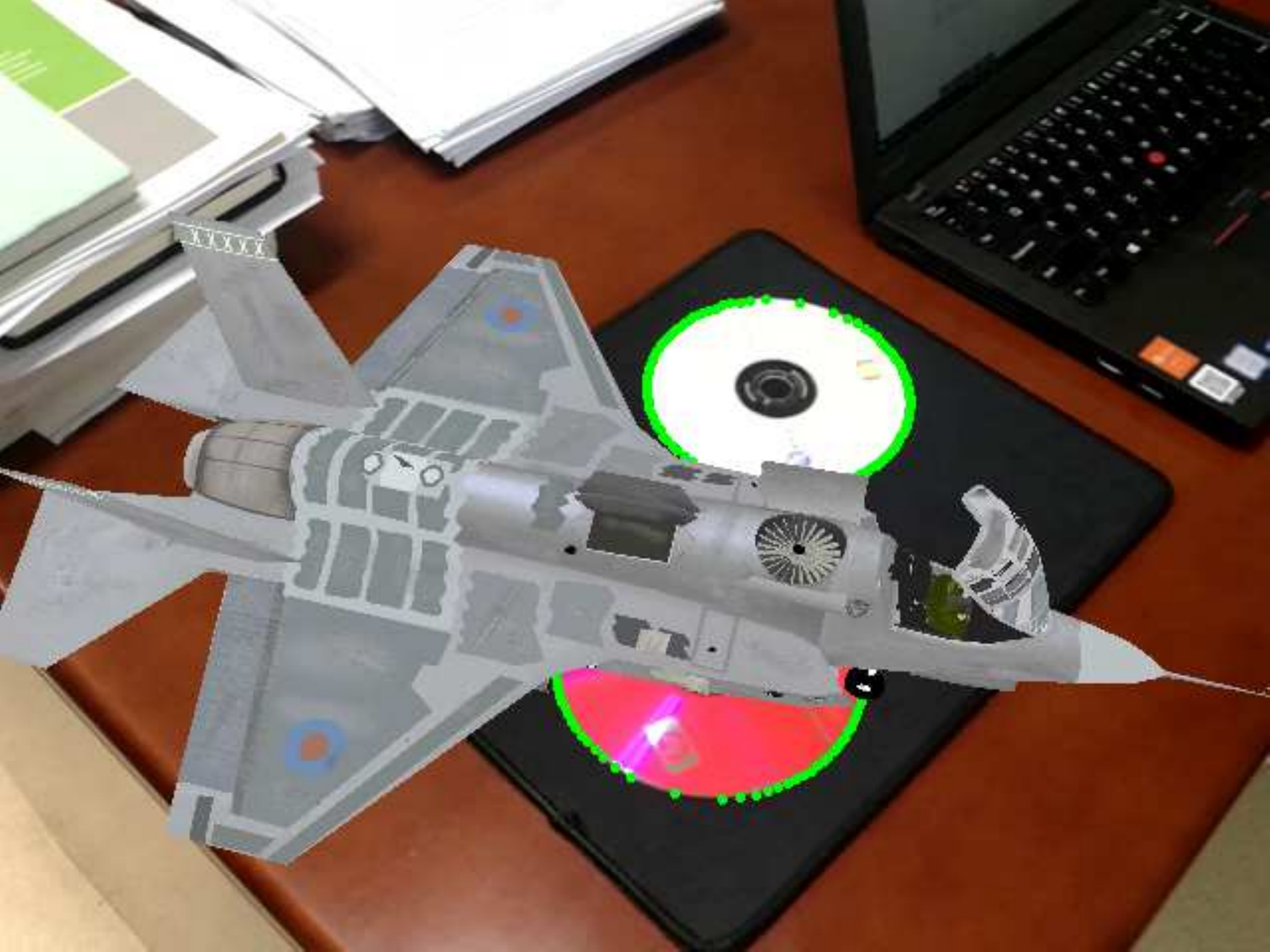}
  \centerline{(c)}
\end{minipage}
\begin{minipage}[b]{0.24\linewidth}
  \centering
  \includegraphics[width=\linewidth]{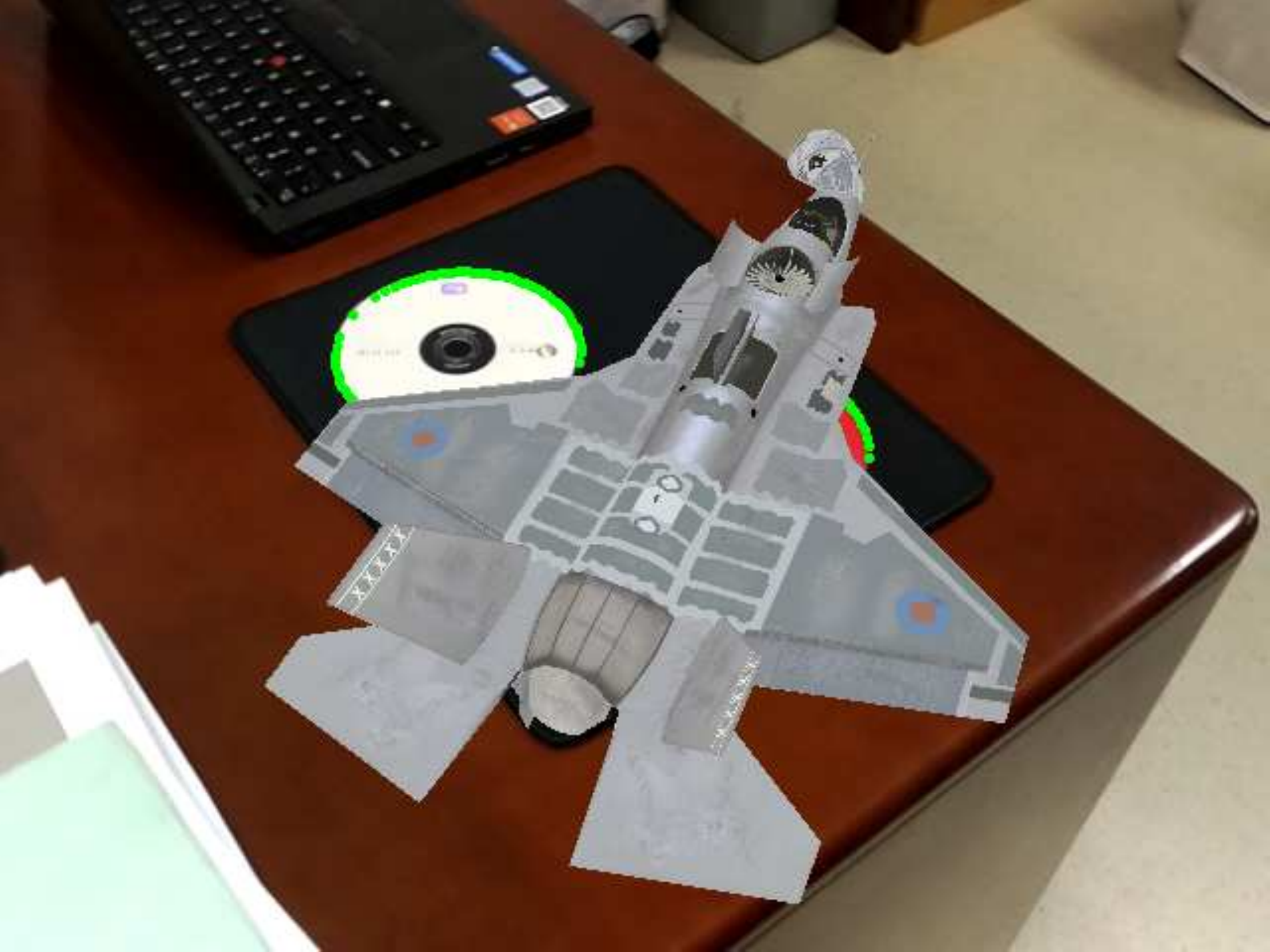}
  \centerline{(d)}
\end{minipage}
\caption{Augmented reality on two videos captured by a rolling shutter camera. (a) and (b) are from a video with weak texture. (c) and (d) are from another video with rich texture.}
\label{fig:long}
\label{fig:onecol}
\end{figure*}

The above experimental results show that the performances of our method outperform ARToolkitPlus, AprilTag2 and RUNETag VS. Gaussian noise, Gaussian blur and distance from camera to marker in terms of localization accuracy and stability. This is mainly due to the used edges, the analytical solutions and the proposed nonlinear optimization, in which no PnP is adopted. For other kinds of markers in Figure 3, there are the similar results. Although we use one point in (b) and (c) of Figure 3 and two points in (d) of Figure 3, identification of them through edges has no confusion and the localization accuracies can be refined further by our proposed nonlinear optimization from the edges.

\subsection{Evaluation on Real Data}
In addition to the evaluation with simulation images, we also perform tests on real videos. Real videos are captured by a digital camera and a mobile phone with a rolling shutter. The image sizes are all 640x480 pixels. Virtual objects are augmented by the computed camera poses to check the stability and accuracy. The used setup is an Intel Xeon(R) E5-1620(eight cores @ 3.50GHz) and 16 GB RAM. In real life, circles are very common. Therefore, markers of our method are easy to be found in real environments. This is different from ARToolkitPlus, AprilTag2 and RUNETag. We use two disks as the marker, a red disk and a white disk.

Figure 9 shows the augmented results with different illuminations, where the airplane is virtual. The scene is brighter in (a) and (b) and darker in (c) and (d). The virtual airplane is overlaid stably by the computed camera pose, although the imaged circles are small. This verifies our method is robust to illumination.
\begin{figure}[t]
\begin{center}
   \includegraphics[width=1.0\linewidth]{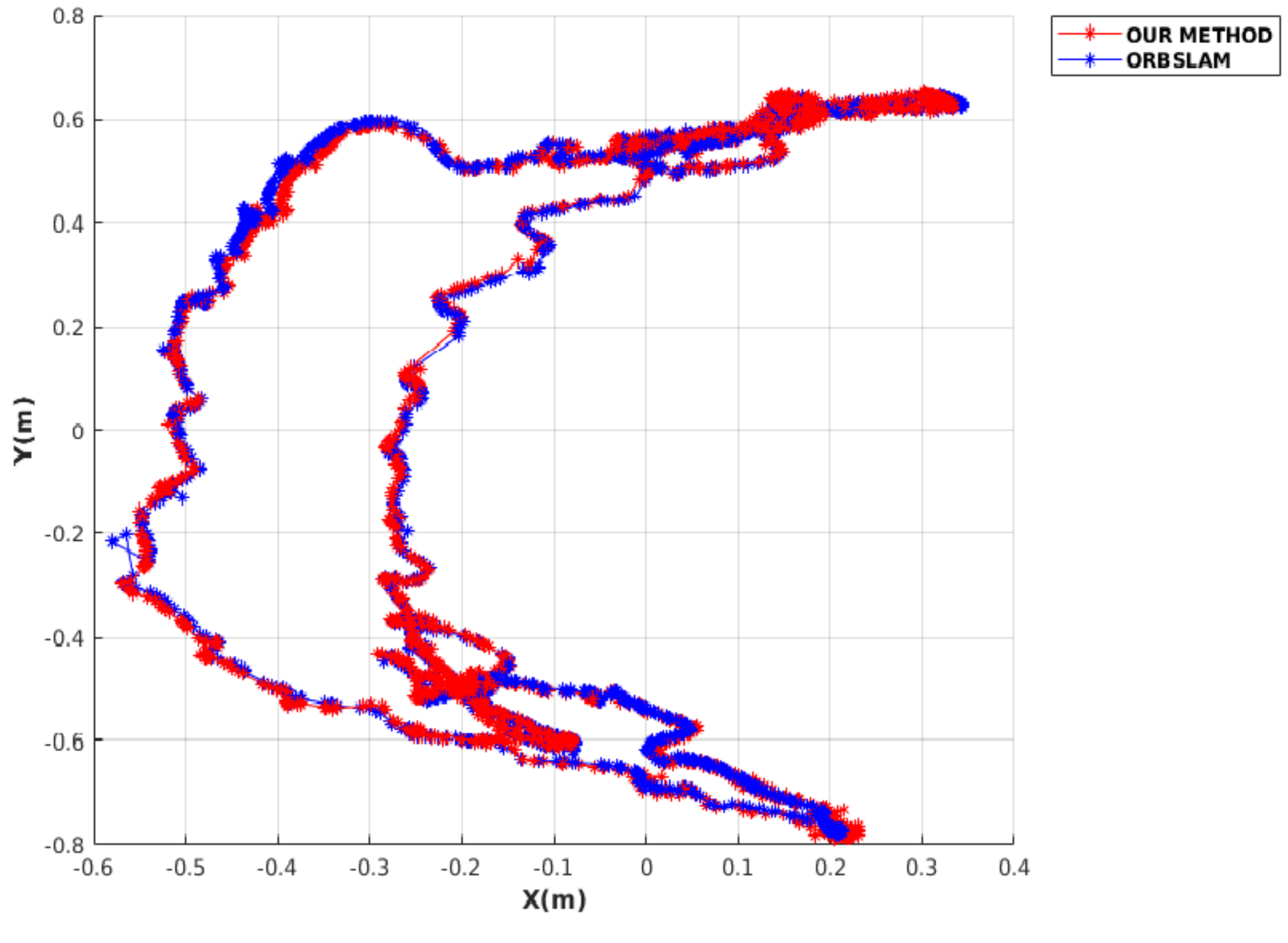}
\end{center}
   \caption{Estimated trajectories of our method and ORBSLAM: the red is by our method and the blue is by ORBSLAM.}
\label{fig:long}
\label{fig:onecol}
\end{figure}
\begin{figure}[t]
\begin{center}
   \includegraphics[width=1.0\linewidth]{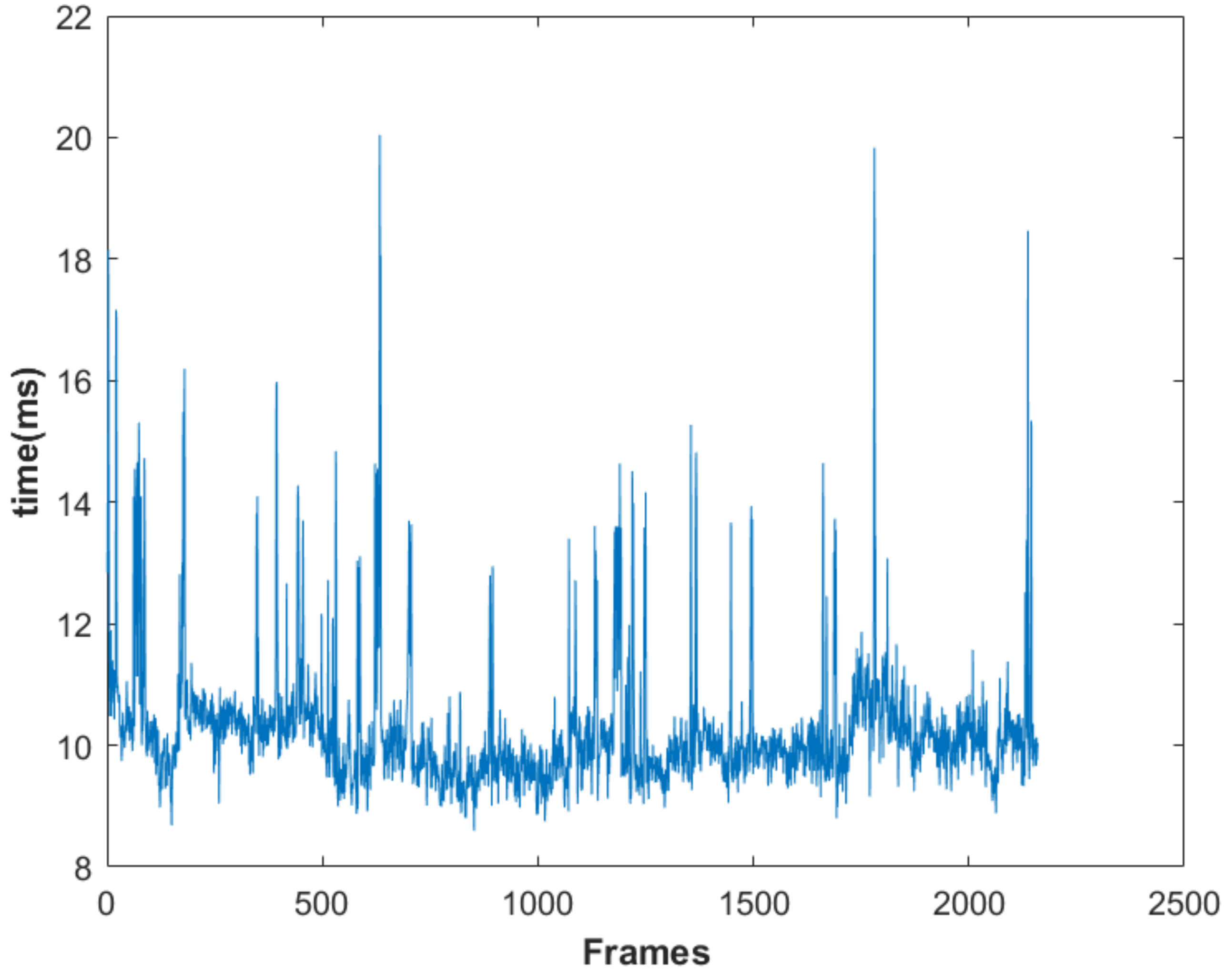}
\end{center}
   \caption{Time consumption of camera localization for each frame, where the average time is about 10ms.}
\label{fig:long}
\label{fig:onecol}
\end{figure}

We also assess our method with real images on different blur levels. Figure 10 shows the augmented results with different image blur in a dark scene, where the airplane is virtual that is placed into the images stably. Although the condition is severe, the performance of our method is still robust.

Images in Figure 9 and Figure 10 above are captured by a digital camera. We further evaluate our method on videos captured by a rolling shutter camera in a mobile phone. Two frames from one video with weak texture are shown as (a) and (b) in Figure 11 and two frames from another video with rich texture are shown as (c) and (d) in Figure 11, where the virtual airplane is augmented stably by our method. Simultaneously, we implement ORBSLAM \cite{mur2015orb} on the same videos of Figure 11. ORBSLAM can not run on the video of (a) and (b) but can on the video of (c) and (d). The camera trajectory of our method and ORBSLAM on the second video are shown in Figure 12. They are similar. But when observing them in detail, we see that the camera trajectory of our method is more smooth than that of ORBSLAM. The mean distance of camera trajectory between our method and ORBSLAM is 1.2cm. We can make a remark that our markers can be put into a real environment to be as an evaluation criterion for general SLAM systems when ground truth can not be obtained easily.

Our method is not only robust but also can run in real time. Figure 13 shows the time consumptions of our method from each frame of a video including 2160 frames, where the time of tracking camera pose is from a minimum of about 8ms to a maximum of about 21ms. The average time of tracking each frame pose is about 10ms. In terms of the speed, our method is very practical for many applications, such as AR and robotics.

\section{Conclusions and Future Work}
A new method for camera pose tracking is proposed from the designed circular markers. The main advantages are: 1)PnP/RANSAC are not needed; 2) The mainly used image features are fitted edges. 3) The given 6D camera pose are analytical and lightweight as very concise forms; 4) A proposed nonlinear cost function based on a geometric distance is used to refine the initial 6D camera pose. These make our camera localization robust, accurate and simultaneously fast. What's more, different from other planar fiducial markers, the markers of our method are easy to be found in real environments. Experimental results show that the proposed method outperforms the state of the arts in term of noise, blur, and distance from camera to marker. At the same time, this method can run at about 100FPS on a consumer computer with only CPU. In the future, the work will be extended to large scale environments like in \cite{munoz2019spm, munoz2018mapping}.

{\small
\bibliographystyle{ieee}
\bibliography{egbib}

\begin{thebibliography}{10}\itemsep=-1pt

\bibitem{ababsa2004robust}
F.-e. Ababsa and M.~Mallem.
\newblock Robust camera pose estimation using 2d fiducials tracking for
  real-time augmented reality systems.
\newblock In {\em Proceedings of the 2004 ACM SIGGRAPH international conference
  on Virtual Reality continuum and its applications in industry}, pages
  431--435. ACM, 2004.

\bibitem{agarwal2012ceres}
S.~Agarwal, K.~Mierle, et~al.
\newblock Ceres solver.
\newblock {\em URL http://ceres-solver.org}, 2012.

\bibitem{bergamasco2016accurate}
F.~Bergamasco, A.~Albarelli, L.~Cosmo, E.~Rodola, and A.~Torsello.
\newblock An accurate and robust artificial marker based on cyclic codes.
\newblock {\em IEEE Trans. Pattern Anal. Mach. Intell.}, 38(12):2359--2373,
  2016.

\bibitem{bergamasco2013pi}
F.~Bergamasco, A.~Albarelli, and A.~Torsello.
\newblock Pi-tag: a fast image-space marker design based on projective
  invariants.
\newblock {\em Machine vision and applications}, 24(6):1295--1310, 2013.

\bibitem{calvet2012camera}
L.~Calvet, P.~Gurdjos, and V.~Charvillat.
\newblock Camera tracking based on circular point factorization.
\newblock In {\em Proceedings of the 21st International Conference on Pattern
  Recognition (ICPR2012)}, pages 2128--2131. IEEE, 2012.

\bibitem{calvet2016detection}
L.~Calvet, P.~Gurdjos, C.~Griwodz, and S.~Gasparini.
\newblock Detection and accurate localization of circular fiducials under
  highly challenging conditions.
\newblock In {\em Proceedings of the IEEE Conference on Computer Vision and
  Pattern Recognition}, pages 562--570, 2016.

\bibitem{chen2004camera}
Q.~Chen, H.~Wu, and T.~Wada.
\newblock Camera calibration with two arbitrary coplanar circles.
\newblock In {\em European Conference on Computer Vision}, pages 521--532.
  Springer, 2004.

\bibitem{claus2005reliable}
D.~Claus and A.~W. Fitzgibbon.
\newblock Reliable automatic calibration of a marker-based position tracking
  system.
\newblock In {\em Application of Computer Vision, 2005. WACV/MOTIONS'05 Volume
  1. Seventh IEEE Workshops on}, volume~1, pages 300--305. IEEE, 2005.

\bibitem{DeGol:ICCV:17}
J.~DeGol, T.~Bretl, and D.~Hoiem.
\newblock Chromatag: A colored marker and fast detection algorithm.
\newblock In {\em ICCV}, 2017.

\bibitem{fiala2005artag}
M.~Fiala.
\newblock Artag, a fiducial marker system using digital techniques.
\newblock In {\em Computer Vision and Pattern Recognition, 2005. CVPR 2005.
  IEEE Computer Society Conference on}, volume~2, pages 590--596. IEEE, 2005.

\bibitem{fiala2010designing}
M.~Fiala.
\newblock Designing highly reliable fiducial markers.
\newblock {\em IEEE Transactions on Pattern analysis and machine intelligence},
  32(7):1317--1324, 2010.

\bibitem{gatrell1992robust}
L.~B. Gatrell, W.~A. Hoff, and C.~W. Sklair.
\newblock Robust image features: Concentric contrasting circles and their image
  extraction.
\newblock In {\em Cooperative Intelligent Robotics in Space II}, volume 1612,
  pages 235--245. International Society for Optics and Photonics, 1992.

\bibitem{kato1999marker}
H.~Kato and M.~Billinghurst.
\newblock Marker tracking and hmd calibration for a video-based augmented
  reality conferencing system.
\newblock In {\em Augmented Reality, 1999.(IWAR'99) Proceedings. 2nd IEEE and
  ACM International Workshop on}, pages 85--94. IEEE, 1999.

\bibitem{knyaz1998development}
V.~A. Knyaz.
\newblock The development of new coded targets for automated point
  identification and non-contact 3d surface measurements.
\newblock {\em IAPRS}, 5:80--85, 1998.

\bibitem{maidi2010performance}
M.~Maidi, J.-Y. Didier, F.~Ababsa, and M.~Mallem.
\newblock A performance study for camera pose estimation using visual marker
  based tracking.
\newblock {\em Machine Vision and Applications}, 21(3):365--376, 2010.

\bibitem{munoz2019spm}
R.~Mu{\~n}oz-Salinas, M.~J. Mar{\'\i}n-Jimenez, and R.~Medina-Carnicer.
\newblock Spm-slam: Simultaneous localization and mapping with squared planar
  markers.
\newblock {\em Pattern Recognition}, 86:156--171, 2019.

\bibitem{munoz2018mapping}
R.~Mu{\~n}oz-Salinas, M.~J. Mar{\'\i}n-Jimenez, E.~Yeguas-Bolivar, and
  R.~Medina-Carnicer.
\newblock Mapping and localization from planar markers.
\newblock {\em Pattern Recognition}, 73:158--171, 2018.

\bibitem{mur2015orb}
R.~Mur-Artal, J.~M.~M. Montiel, and J.~D. Tardos.
\newblock Orb-slam: a versatile and accurate monocular slam system.
\newblock {\em IEEE Transactions on Robotics}, 31(5):1147--1163, 2015.

\bibitem{mur2017orb}
R.~Mur-Artal and J.~D. Tard{\'o}s.
\newblock Orb-slam2: An open-source slam system for monocular, stereo, and
  rgb-d cameras.
\newblock {\em IEEE Transactions on Robotics}, 33(5):1255--1262, 2017.

\bibitem{naimark2002circular}
L.~Naimark and E.~Foxlin.
\newblock Circular data matrix fiducial system and robust image processing for
  a wearable vision-inertial self-tracker.
\newblock In {\em Proceedings of the 1st International Symposium on Mixed and
  Augmented Reality}, page~27. IEEE Computer Society, 2002.

\bibitem{neumannmulti}
Y.~C. J. L.~U. Neumann.
\newblock A multi-ring color fiducial system and a rule-based detection method
  for scalable fiducial-tracking augmented reality.
\newblock {\em Proceedings of International Workshop on Augmented Reality},
  1998.

\bibitem{olson2011apriltag}
E.~Olson.
\newblock Apriltag: A robust and flexible visual fiducial system.
\newblock In {\em Robotics and Automation (ICRA), 2011 IEEE International
  Conference on}, pages 3400--3407. IEEE, 2011.

\bibitem{prasad2015motion}
M.~G. Prasad, S.~Chandran, and M.~S. Brown.
\newblock A motion blur resilient fiducial for quadcopter imaging.
\newblock In {\em Applications of Computer Vision (WACV), 2015 IEEE Winter
  Conference on}, pages 254--261. IEEE, 2015.

\bibitem{toyoura2014mono}
M.~Toyoura, H.~Aruga, M.~Turk, and X.~Mao.
\newblock Mono-spectrum marker: an ar marker robust to image blur and defocus.
\newblock {\em The Visual Computer}, 30(9):1035--1044, 2014.

\bibitem{wagner2010real}
D.~Wagner, G.~Reitmayr, A.~Mulloni, T.~Drummond, and D.~Schmalstieg.
\newblock Real-time detection and tracking for augmented reality on mobile
  phones.
\newblock {\em IEEE transactions on visualization and computer graphics},
  16(3):355--368, 2010.

\bibitem{wagner2007artoolkitplus}
D.~Wagner and D.~Schmalstieg.
\newblock Artoolkitplus for pose tracking on mobile devices.
\newblock {\em Proceedings of 12th Computer Vision Winter Workshop}, pages
  139--146, 2007.

\bibitem{wang2016apriltag}
J.~Wang and E.~Olson.
\newblock Apriltag 2: Efficient and robust fiducial detection.
\newblock In {\em IROS}, pages 4193--4198, 2016.

\bibitem{wu2019efficient}
Y.~Wu, H.~Wang, F.~Tang, and Z.~Wang.
\newblock Efficient conic fitting with an analytical polar-n-direction
  geometric distance.
\newblock {\em Pattern Recognition}, 90:415--423, 2019.

\bibitem{wu2004camera}
Y.~Wu, H.~Zhu, Z.~Hu, and F.~Wu.
\newblock Camera calibration from the quasi-affine invariance of two parallel
  circles.
\newblock In {\em European Conference on Computer Vision}, pages 190--202.
  Springer, 2004.

\end{thebibliography}
}

\end{document}